\documentclass[10pt,twocolumn,letterpaper]{article}

\usepackage{cvpr}
\usepackage{times}
\usepackage{epsfig}
\usepackage{graphicx}
\usepackage{amsmath}
\usepackage{amssymb}
\usepackage{multirow}
\usepackage{color}

\usepackage[pagebackref=true,breaklinks=true,letterpaper=true,colorlinks,bookmarks=false]{hyperref}

\newcommand{\umls}{UMLS\textsuperscript{\textregistered}}

\cvprfinalcopy 

\def\cvprPaperID{766} 

\ifcvprfinal\fi
\begin{document}

\renewcommand{\dblfloatpagefraction}{1}
\renewcommand{\textfraction}{0}
\renewcommand{\dblfloatsep}{5pt}
\renewcommand{\dbltextfloatsep}{1ex}
\renewcommand{\dbltopfraction}{1}
\renewcommand{\intextsep}{5pt}
\renewcommand{\floatpagefraction}{1}
\renewcommand{\floatsep}{5pt}
\renewcommand{\textfloatsep}{1ex}
\renewcommand{\topfraction}{1}
\renewcommand{\abovecaptionskip}{2pt}
\let\OldCaption=\caption
\renewcommand{\caption}[1]{\small\OldCaption{\em#1}}

\makeatletter
\def\@normalsize{\@setsize\normalsize{10pt}\xpt\@xpt
\abovedisplayskip 10pt plus2pt minus5pt\belowdisplayskip
\abovedisplayskip \abovedisplayshortskip \z@
plus3pt\belowdisplayshortskip 6pt plus3pt
minus3pt\let\@listi\@listI}
\def\subsize{\@setsize\subsize{12pt}\xipt\@xipt}
\def\section{\@startsection {section}{1}{\z@}{1.0ex plus
1ex minus .2ex}{.2ex plus .2ex}{\large\bf}}
\def\subsection{\@startsection {subsection}{2}{\z@}{.2ex
plus 1ex} {.2ex plus .2ex}{\subsize\bf}} \makeatother

\newcommand{\Section}[1]{\section{\hskip -1em.~~#1}}
\newcommand{\SubSection}[1]{\subsection{\hskip -1em.~~#1}}
\def\@listI{%
 \leftmargin\leftmargini
 \partopsep 0pt
 \parsep 0pt
 \topsep 0pt
 \itemsep pt
 \relax
} \long\def\@makecaption#1#2{
 \vskip -5pt
 \setbox\@tempboxa\hbox{\small{#1\,:\,#2}}
  \ifdim \wd\@tempboxa >\hsize \unhbox\@tempboxa\par \else
  \hbox to\hsize{\hfil\box\@tempboxa\hfil}
\fi \vskip -0.2cm}

\jot=0pt \abovedisplayskip=3pt \belowdisplayskip=3pt
\abovedisplayshortskip=0pt \belowdisplayshortskip=0pt

\title{ChestX-ray8: Hospital-scale Chest X-ray Database and Benchmarks on \\ Weakly-Supervised Classification and Localization of Common Thorax Diseases}

\author{Xiaosong Wang$^1$, Yifan Peng $^2$, Le Lu $^1$, Zhiyong Lu $^2$, Mohammadhadi Bagheri $^1$, Ronald M. Summers $^1$ \\
$^1$Department of Radiology and Imaging Sciences, Clinical Center,\\
$^2$ National Center for Biotechnology Information, National Library of Medicine, \\
National Institutes of Health, Bethesda, MD 20892\\
{\tt\small \{xiaosong.wang,yifan.peng,le.lu,luzh,mohammad.bagheri,rms\}@nih.gov}
}


\maketitle


\begin{abstract}
The chest X-ray is one of the most commonly accessible radiological examinations for screening and diagnosis of many lung diseases. A tremendous number of X-ray imaging studies accompanied by radiological reports are accumulated and stored in many modern hospitals' Picture Archiving and Communication Systems (PACS). On the other side, it is still an open question how this type of hospital-size knowledge database containing invaluable imaging informatics (i.e., loosely labeled) can be used to facilitate the data-hungry deep learning paradigms in building truly large-scale high precision computer-aided diagnosis (CAD) systems. 

In this paper, we present a new chest X-ray database, namely ``ChestX-ray8", which comprises 108,948 frontal-view X-ray images of 32,717 unique patients with the text-mined eight disease image labels (where each image can have multi-labels), from the associated radiological reports using natural language processing. Importantly, we demonstrate that these commonly occurring thoracic diseases can be detected and even spatially-located via a unified weakly-supervised multi-label image classification and disease localization framework, which is validated using our proposed dataset. 
Although the initial quantitative results are promising as reported, deep convolutional neural network based ``reading chest X-rays" (i.e., recognizing and locating the common disease patterns trained with only image-level labels) remains a strenuous task for fully-automated high precision CAD systems. 
\end{abstract} 

\begin{figure}[]
	\includegraphics[width=1\columnwidth]{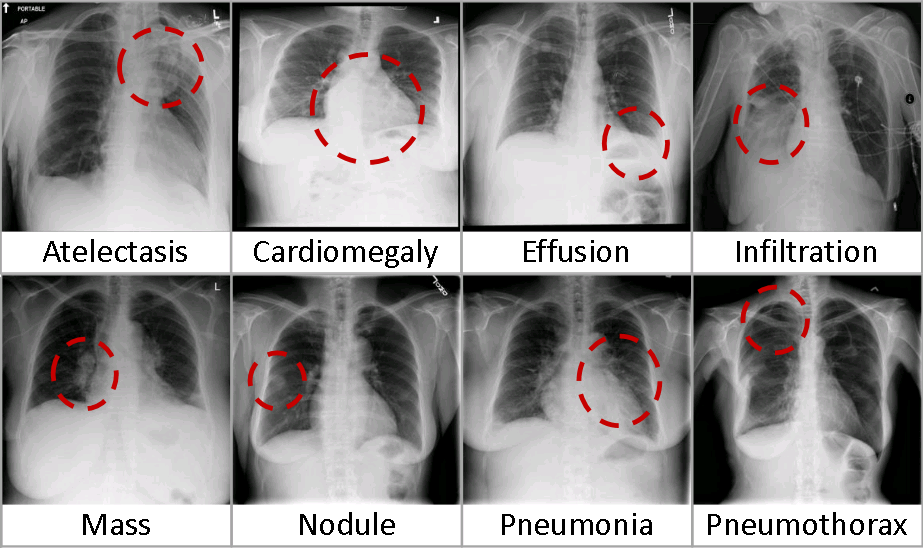}
	\caption{Eight common thoracic diseases observed in chest X-rays that validate a challenging task of fully-automated diagnosis.}
	\label{fig:8_diseases}
\end{figure}

\section{Introduction}
The rapid and tremendous progress has been evidenced in a range of computer vision problems via deep learning and large-scale annotated image datasets \cite{krizhevsky2012imagenet,Russakovsky2015,Everingham2015,Lin2014COCO}. Drastically improved quantitative performances in object recognition, detection and segmentation
are demonstrated in comparison to previous shallow methodologies built upon hand-crafted image features. Deep neural network representations further make the joint language and vision learning tasks more feasible to solve, in image captioning \cite{Vinyals2015Show,Karpathy2015Deep,Plummer2015Flickr30k,Vendrov2016Order,Johnson2016DenseCap}, visual question answering \cite{Antol2015VQA,Tapaswi2015MovieQA,Wu2016Ask,Zhu2016Visual7W} and knowledge-guided transfer learning \cite{Ba2015Predicting,Qiao2016Less}, and so on. However, the intriguing and strongly observable performance gaps of the current state-of-the-art object detection and segmentation methods, evaluated between using PASCAL VOC \cite{Everingham2015} and employing Microsoft (MS) COCO \cite{Lin2014COCO}, demonstrate that there is still significant  room for performance improvement when underlying challenges (represented by different datasets) become greater. For example, MS COCO is composed of 80 object categories from 200k images, with 1.2M instances (350k are people) where every instance is segmented and many instances are small objects. Comparing to PASCAL VOC of only 20 classes and 11,530 images containing 27,450 annotated objects with bounding-boxes (BBox), the top competing object detection approaches achieve in 0.413 in MS COCO 
versus 0.884 in PASCAL VOC 
under mean Average Precision (mAP). 

Deep learning yields similar rises in performance in the medical image analysis domain for object (often human anatomical or pathological structures in radiology imaging) detection and segmentation tasks. Recent notable work includes (but do not limit to) an overview review on the future promise of deep learning \cite{Greenspan2016Guest} and a collection of important medical applications on lymph node and interstitial lung disease detection and classification \cite{roth2014new,Shin2016Deep}; cerebral microbleed detection \cite{Dou2016Automatic}; pulmonary nodule detection in CT images \cite{Setio2016Pulmonary}; automated pancreas segmentation \cite{roth2015deeporgan}; cell image segmentation and tracking \cite{ronneberger2015unet}, predicting spinal radiological scores \cite{Jamaludin2016SpineNet} and extensions of multi-modal imaging segmentation \cite{Moeskops2016Deep,Havaei2016HeMIS}. The main limitation is that all proposed methods are evaluated on some small-to-middle scale problems of (at most) several hundred patients. It remains unclear how well the current deep learning techniques will scale up to tens of thousands of patient studies.

In the era of deep learning in computer vision, research efforts on building various annotated image datasets \cite{Russakovsky2015,Everingham2015,Lin2014COCO,Antol2015VQA,Plummer2015Flickr30k,Zhu2016Visual7W,Johnson2016DenseCap,krishnavisualgenome} with different characteristics play indispensably important roles on the better definition of the forthcoming problems, challenges and subsequently possible technological progresses. Particularly, here we focus on the relationship and joint learning of image (chest X-rays) and text (X-ray reports). The previous representative image caption generation work \cite{Vinyals2015Show,Karpathy2015Deep} utilize Flickr8K, Flickr30K \cite{Young2014From} and MS COCO \cite{Lin2014COCO} datasets that hold 8,000, 31,000 and 123,000 images respectively and every image is annotated by five sentences via Amazon Mechanical Turk (AMT). The text generally describes annotator's attention of objects and activity occurring on an image in a straightforward manner. Region-level ImageNet pre-trained convolutional neural networks (CNN) based detectors are used to parse an input image and output a list of attributes or ``visually-grounded high-level concepts'' (including objects, actions, scenes and so on) in \cite{Karpathy2015Deep,Wu2016Ask}. Visual question answering (VQA) requires more detailed parsing and complex reasoning on the image contents to answer the paired natural language questions. A new dataset containing 250k natural images, 760k questions and 10M text answers \cite{Antol2015VQA} is provided to address this new challenge. Additionally, databases such as ``Flickr30k Entities'' \cite{Plummer2015Flickr30k}, ``Visual7W'' \cite{Zhu2016Visual7W} and ``Visual Genome'' \cite{krishnavisualgenome,Johnson2016DenseCap} (as detailed as 94,000 images and 4,100,000 region-grounded captions) are introduced to construct and learn the spatially-dense and increasingly difficult semantic links between textual descriptions and image regions through the object-level grounding. 

Though one could argue that the high-level analogy exists between image caption generation, visual question answering and imaging based disease diagnosis \cite{Shin2016Learning,Shin2016Interleaved}, there are three factors making truly large-scale medical image based diagnosis (e.g., involving tens of thousands of patients) tremendously more formidable. {\bf 1}, Generic, open-ended image-level anatomy and pathology labels cannot be obtained through crowd-sourcing, such as AMT, which is prohibitively implausible for non-medically trained annotators. Therefore we exploit to mine the per-image (possibly multiple) common thoracic pathology labels from the image-attached chest X-ray radiological reports using Natural Language Processing (NLP) techniques. Radiologists tend to write more abstract and complex logical reasoning sentences than the plain describing texts in \cite{Young2014From,Lin2014COCO}. {\bf 2}, The spatial dimensions of an chest X-ray are usually $2000 \times 3000$ pixels. Local pathological image regions can show hugely varying sizes or extents but often very small comparing to the full image scale. Fig. \ref{fig:8_diseases} shows eight illustrative examples and the actual pathological findings are often significantly smaller (thus harder to detect). Fully dense annotation of region-level bounding boxes (for grounding the pathological findings) would normally be needed in computer vision datasets \cite{Plummer2015Flickr30k,Zhu2016Visual7W,krishnavisualgenome} but may be completely nonviable for the time being. Consequently, we formulate and verify a weakly-supervised multi-label image classification and disease localization framework to address this difficulty. {\bf 3}, So far, all image captioning and VQA techniques in computer vision strongly depend on the ImageNet pre-trained deep CNN models which already perform very well in a large number of object classes and serves a good baseline for further model fine-tuning. However, this situation does not apply to the medical image diagnosis domain. Thus we have to learn the deep image recognition and localization models while constructing the weakly-labeled medical image database. 

To tackle these issues, we propose a new chest X-ray database, namely ``ChestX-ray8'', which comprises 108,948 frontal-view X-ray images of 32,717 (collected from the year of 1992 to 2015) unique patients with the text-mined eight common disease labels, mined from the text radiological reports via NLP techniques. In particular, we demonstrate that these commonly occurred thoracic diseases can be detected and even spatially-located via a unified weakly-supervised multi-label image classification and disease localization formulation. Our initial quantitative results are promising. However developing fully-automated deep learning based ``reading chest X-rays" systems is still an arduous journey to be exploited. Details of accessing the ChestX-ray8 dataset can be found via the website \footnote{ 
\url{https://nihcc.app.box.com/v/ChestXray-NIHCC}, more details: \url{https://www.cc.nih.gov/drd/summers.html}}.

\subsection{Related Work}

There have been recent efforts on creating openly available annotated medical image databases \cite{Jamaludin2016,Yao2016,roth2014new,roth2015deeporgan} with the studied patient numbers ranging from a few hundreds to two thousands. Particularly for chest X-rays, the largest public dataset is OpenI \cite{openi} that contains
3,955 radiology reports from the Indiana Network for Patient Care and 7,470 associated chest x-rays from the hospitals’ picture archiving and communication system (PACS). This database is utilized in \cite{Shin2016Learning} as a problem of caption generation but no quantitative disease detection results are reported. Our newly proposed chest X-ray database is at least one order of magnitude larger than OpenI \cite{openi} (Refer to Table \ref{tab:openi_corpus}). To achieve the better clinical relevance, we focus to exploit the quantitative performance on weakly-supervised multi-label image classification and disease localization of common thoracic diseases, in analogy to the intermediate step of ``detecting attributes'' in \cite{Wu2016Ask} or ``visual grounding'' for \cite{Plummer2015Flickr30k,Zhu2016Visual7W,Johnson2016DenseCap}.


\section{Construction of Hospital-scale Chest X-ray Database}

\begin{figure}[t]
	\centering
	\includegraphics[width=1\columnwidth]{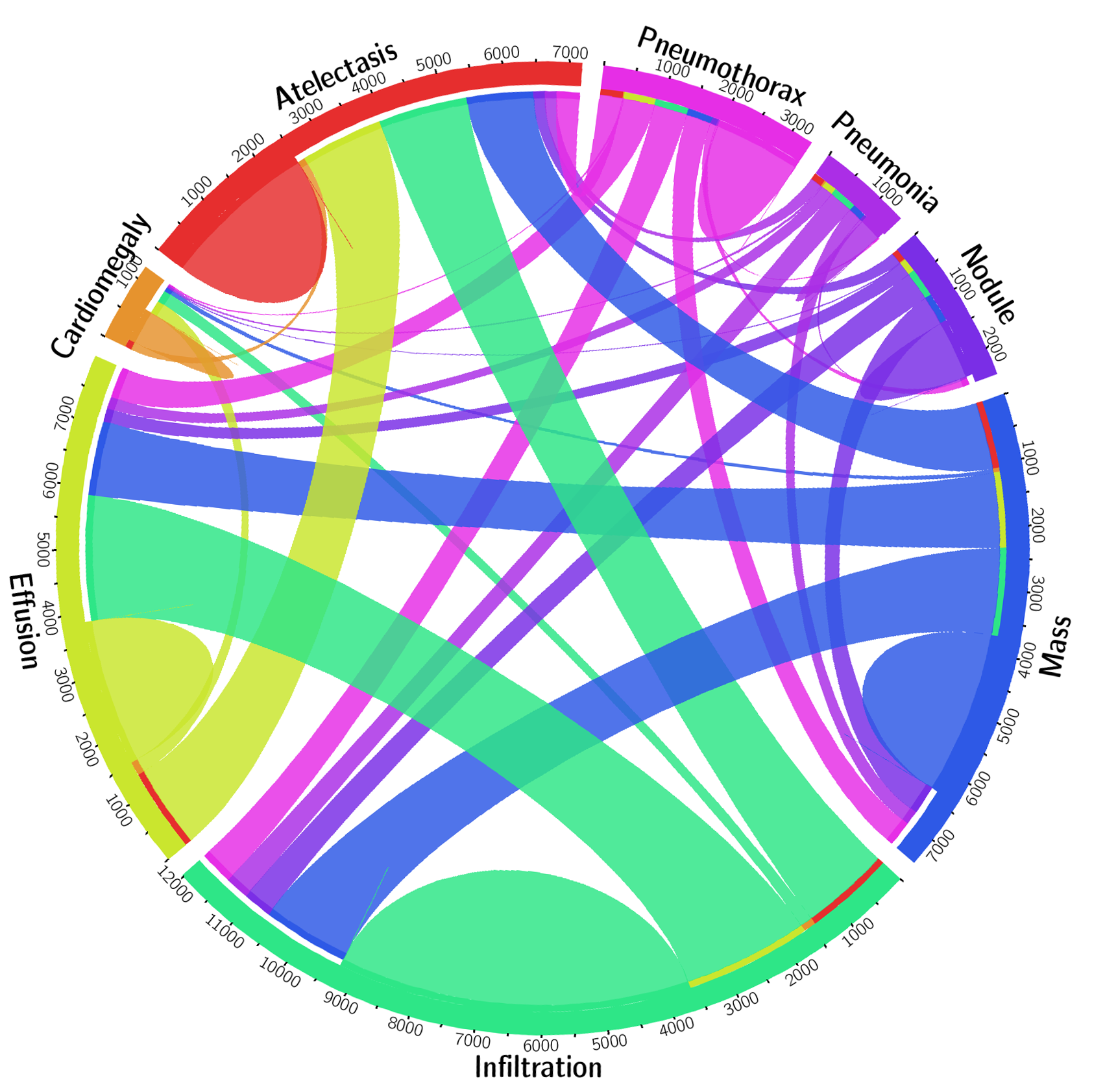}
	\caption{The circular diagram shows the proportions of images with multi-labels in each of 8 pathology classes and the labels' co-occurrence statistics.}
	\label{fig:keywords_statistics}
\end{figure}

In this section, we describe the approach for building a hospital-scale chest X-ray image database, namely ``ChestX-ray8'', mined from our institute's PACS system. First, we short-list eight common thoracic pathology keywords that are frequently observed and diagnosed, i.e., Atelectasis, Cardiomegaly, Effusion, Infiltration, Mass, Nodule, Pneumonia and Pneumathorax (Fig. \ref{fig:8_diseases}), based on radiologists' feedback. Given those 8 text keywords, we search the PACS system to pull out all the related radiological reports (together with images) as our target corpus. A variety of Natural Language Processing (NLP) techniques are adopted for detecting the pathology keywords and removal of negation and uncertainty. Each radiological report will be either linked with one or more keywords or marked with 'Normal' as the background category. As a result, the ChestX-ray8 database is composed of 108,948 frontal-view X-ray images (from 32,717 patients) and each image is labeled with one or multiple pathology keywords or ``Normal'' otherwise. Fig.~\ref{fig:keywords_statistics} illustrates the correlation of the resulted keywords. It reveals some connections between different pathologies, which agree with radiologists' domain knowledge, e.g., Infiltration is often associated with Atelectasis and Effusion. To some extend, this is similar with understanding the interactions and relationships among objects or concepts in natural images \cite{krishnavisualgenome}. 

\subsection{Labeling Disease Names by Text Mining}


Overall, our approach produces labels using the reports in two passes. In the first iteration, we detected all the disease concept in the corpus. The main body of each chest X-ray report is generally structured as ``Comparison", ``Indication", ``Findings", and ``Impression'' sections. Here, we focus on detecting disease concepts in the Findings and Impression sections. If a report contains neither of these two sections, the full-length report will then be considered. In the second pass, we code the reports as ``Normal'' if they do not contain any diseases (not limited to 8 predefined pathologies). 

{\bf Pathology Detection:} 
We mine the radiology reports for disease concepts using two tools, DNorm~\cite{leaman2015challenges} and MetaMap~\cite{aronson2010overview}. DNorm is a machine learning method for disease recognition and normalization. It maps every mention of keywords in a report to a unique concept ID in the Systematized Nomenclature of Medicine – Clinical Terms (or SNOMED-CT), which is a standardized vocabulary of clinical terminology for the electronic exchange of clinical health information.

MetaMap is another prominent tool to detect bio-concepts from the biomedical text corpus. Different from DNorm, it is an ontology-based approach for the detection of Unified Medical Language System\textsuperscript{\textregistered} (\umls) Metathesaurus. In this work, we only consider the semantic types of Diseases or Syndromes and Findings (namely `dsyn' and `fndg' respectively). To maximize the recall of our automatic disease detection, we merge the results of DNorm and MetaMap. Table 1 (in the supplementary material) shows the corresponding SNOMED-CT concepts that are relevant to the eight target diseases (these mappings are developed by searching the disease names in the~\umls terminology service~\footnote{\url{https://uts.nlm.nih.gov/metathesaurus.html}}, and verified by a board-certified radiologist.


{\bf Negation and Uncertainty:}
The disease detection algorithm locates every keyword mentioned in the radiology report no matter if it is truly present or negated. To eliminate the noisy labeling, we need to rule out those negated pathological statements and, more importantly, uncertain mentions of findings and diseases, e.g., ``suggesting obstructive lung disease''.

\begin{figure}
	\includegraphics[width=\columnwidth]{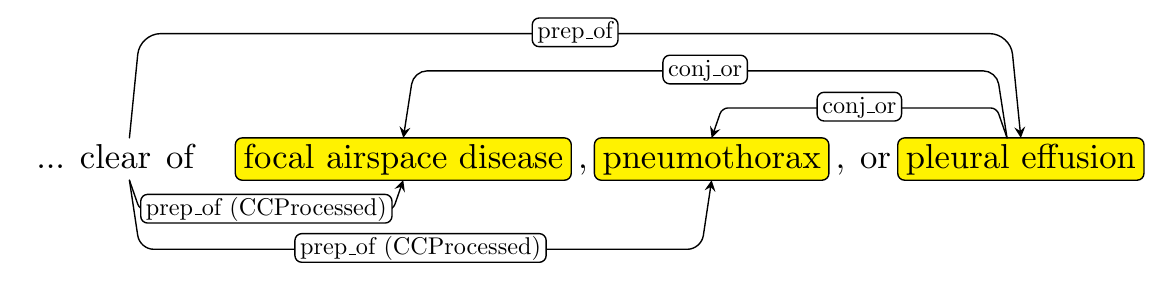}
	\caption{The dependency graph of text: ``clear of focal airspace disease, pneumothorax, or pleural effusion''.}
	\label{fig:dependency_graph}
\end{figure}

Although many text processing systems (such as \cite{chapman2001simple}) can handle the negation/uncertainty detection problem, most of them exploit regular expressions on the text directly. One of the disadvantages to use regular expressions for negation/uncertainty detection is that they cannot capture various syntactic constructions for multiple subjects. For example, in the phrase of ``clear of A and B'', the regular expression can capture ``A'' as a negation but not ``B'', particularly when both ``A'' and ``B'' are long and complex noun phrases (``clear of focal airspace disease, pneumothorax, or pleural effusion'' in Fig. \ref{fig:dependency_graph}). 

To overcome this complication, we hand-craft a number of novel rules of negation/uncertainty defined on the syntactic level in this work. More specifically, we utilize the syntactic dependency information because it is close to the semantic relationship between words and thus has become prevalent in biomedical text processing. We defined our rules on the dependency graph, by utilizing the dependency label and direction information between words. 

As the first step of preprocessing, we split and tokenize the reports into sentences using NLTK~\cite{bird2009natural}. Next we parse each sentence by the Bllip parser~\cite{charniak2005coarse} using David McClosky’s biomedical model~\cite{mcclosky2009any}. The syntactic dependencies are then obtained from ``CCProcessed'' dependencies output by applying Stanford dependencies converter~\cite{demarneffe2015stanford} on the parse tree. The ``CCProcessed'' representation propagates conjunct dependencies thus simplifies coordinations. As a result, we can use fewer rules to match more complex constructions. For an example as shown in Fig.~\ref{fig:dependency_graph}, we could use ``clear $\rightarrow$ prep\_of $\rightarrow$ DISEASE'' to detect three negations from the text $\langle$neg, focal airspace disease$\rangle$, $\langle$neg, pneumothorax$\rangle$, and $\langle$neg, pleural effusion$\rangle$.

Furthermore, we label a radiology report as ``normal'' if it meets one of the following criteria:
%
\begin{itemize}
	\item If there is no disease detected in the report. Note that here we not only consider 8 diseases of interest in this paper, but all diseases detected in the reports.
	\item If the report contains text-mined concepts of ``normal'' or ``normal size'' (CUIs C0205307 and C0332506  in the SNOMED-CT concepts respectively).
\end{itemize}

\subsection{Quality Control on Disease Labeling}

To validate our method, we perform the following experiments. Given the fact that no gold-standard labels exist for our dataset, we resort to some existing annotated corpora as an alternative. Using the OpenI API 
\cite{openi}, we retrieve a total of 3,851 unique radiology reports where each OpenI report is assigned with its key findings/disease names by human annotators~\cite{demner2015preparing}. Given our focus on the eight diseases, a subset of OpenI reports and their human annotations are used as the gold standard for evaluating our method. Table~\ref{tab:openi_corpus} summarizes the statistics of the subset of OpenI~\cite{openi,QIMS5132} reports. 
\begin{table}
	\begin{center}
		\begin{tabular}{l|r|r|r|r}
			\hline
			Item \# & OpenI & Ov. & ChestX-ray8 & Ov.\\
			\hline\hline
			Report & 2,435 & - & 108,948 & -\\
			Annotations & 2,435 & - & - & -\\
			Atelectasis & 315 & 122 & 5,789 & 3,286 \\
			Cardiomegaly & 345 & 100 & 1,010 & 475\\
			Effusion & 153 & 94 & 6,331 & 4,017\\
			Infiltration & 60 & 45 & 10,317 & 4,698\\
			Mass & 15 & 4 & 6,046 & 3,432\\
			Nodule & 106 & 18 & 1,971 & 1,041 \\
			Pneumonia & 40 & 15 & 1,062 & 703\\
			Pneumothorax & 22 & 11 & 2,793 & 1,403\\
			Normal & 1,379 & 0 & 84,312 & 0\\
			\hline
		\end{tabular}
	\end{center}
	\caption{Total number (\#) and \# of Overlap (Ov.) of the corpus in both OpenI and ChestX-ray8 datasets.}
	\label{tab:openi_corpus}
\end{table}
Table~\ref{tab:evaluation of openi} shows the results of our method using OpenI, measured in precision (P), recall (R), and F1-score. Higher precision of 0.90, higher recall of 0.91, and higher F1-score of 0.90 are achieved compared to the existing MetaMap approach (with NegEx enabled). For all diseases, our method obtains higher precisions, particularly in ``pneumothorax'' (0.90 vs. 0.32) and ``infiltration'' (0.74 vs. 0.25). This indicates that the usage of negation and uncertainty detection on syntactic level successfully removes false positive cases. More importantly, the higher precisions meet our expectation to generate a Chest X-ray corpus with accurate semantic labels, to lay a solid foundation for the later processes. \vspace{-3mm}
\begin{table}
	\begin{center}
		\begin{tabular}{lr@{~/~}r@{~/~}rr@{~/~}r@{~/~}r}
			\hline
			\multirow{2}{*}{Disease} & \multicolumn{3}{c}{MetaMap} & \multicolumn{3}{c}{Our Method}\\\cline{2-7}
			& P & R & F & P & R & F\\\hline\hline
			Atelectasis & 0.95 & 0.95 & 0.95 & 0.99 & 0.85 & 0.91\\
			Cardiomegaly & 0.99 & 0.83 & 0.90 & 1.00 & 0.79 & 0.88\\
			Effusion & 0.74 & 0.90 & 0.81 & 0.93 & 0.82 & 0.87\\
			Infiltration & 0.25 & 0.98 & 0.39 & 0.74 & 0.87 & 0.80\\
			Mass & 0.59 & 0.67 & 0.62 & 0.75 & 0.40 & 0.52\\
			Nodule & 0.95 & 0.65 & 0.77 & 0.96 & 0.62 & 0.75\\
			Normal & 0.93 & 0.90 & 0.91 & 0.87 & 0.99 & 0.93\\
			Pneumonia & 0.58 & 0.93 & 0.71 & 0.66 & 0.93 & 0.77\\
			Pneumothorax & 0.32 & 0.82 & 0.46 & 0.90 & 0.82 & 0.86\\
			\hspace*{1em}\textit{Total} & 0.84 & 0.88 & 0.86 & 0.90 & 0.91 & 0.90\\
			\hline
		\end{tabular}
	\end{center}
	\caption{Evaluation of image labeling results on OpenI dataset. Performance is reported using P, R, F1-score.}
	\label{tab:evaluation of openi}
\end{table}

\subsection{Processing Chest X-ray Images}
Comparing to the popular ImageNet classification problem, significantly smaller spatial extents of many diseases inside the typical X-ray image dimensions of $3000 \times 2000$ pixels impose challenges in both the capacity of computing hardware and the design of deep learning paradigm. In ChestX-ray8, X-rays images are directly extracted from the DICOM file and resized as $1024\times 1024$ bitmap images without significantly losing the detail contents, compared with image sizes of $512\times 512$ in OpenI dataset. Their intensity ranges are rescaled using the default window settings stored in the DICOM header files. 

\subsection{Bounding Box for Pathologies}
As part of the ChestX-ray8 database, a small number of images with pathology are provided with hand labeled bounding boxes (B-Boxes), which can be used as the ground truth to evaluate the disease localization performance. Furthermore, it could also be adopted for one/low-shot learning setup \cite{hariharan2016low}, in which only one or several samples are needed to initialize the learning and the system will then evolve by itself with more unlabeled data. We leave this as future work.

In our labeling process, we first select 200 instances for each pathology (1,600 instances total), consisting of 983 images. Given an image and a disease keyword, a board-certified radiologist identified only the corresponding disease instance in the image and labeled it with a B-Box. The B-Box is then outputted as an XML file. If one image contains multiple disease instances, each disease instance is labeled separately and stored into individual XML files. As an application of the proposed ChestX-ray8 database and benchmarking, we will demonstrate the detection and localization of thoracic diseases in the following.


\section{Common Thoracic Disease Detection and Localization}

Reading and diagnosing Chest X-ray images may be an entry-level task for radiologists but, in fact it is a complex reasoning problem which often requires careful observation and good knowledge of anatomical principles, physiology and pathology. Such factors increase the difficulty of developing a consistent and automated technique for reading chest X-ray images while simultaneously considering all common thoracic diseases. 

As the main application of ChestX-ray8 dataset, we present a unified weakly-supervised multi-label image classification and pathology localization framework, which can detect the presence of multiple pathologies and subsequently generate bounding boxes around the corresponding pathologies. In details, we tailor Deep Convolutional Neural Network (DCNN) architectures for weakly-supervised object localization, by considering large image capacity, various multi-label CNN losses and different pooling strategies.

\subsection{Unified DCNN Framework}
\begin{figure*}[t]
	\includegraphics[width=1.0\linewidth]{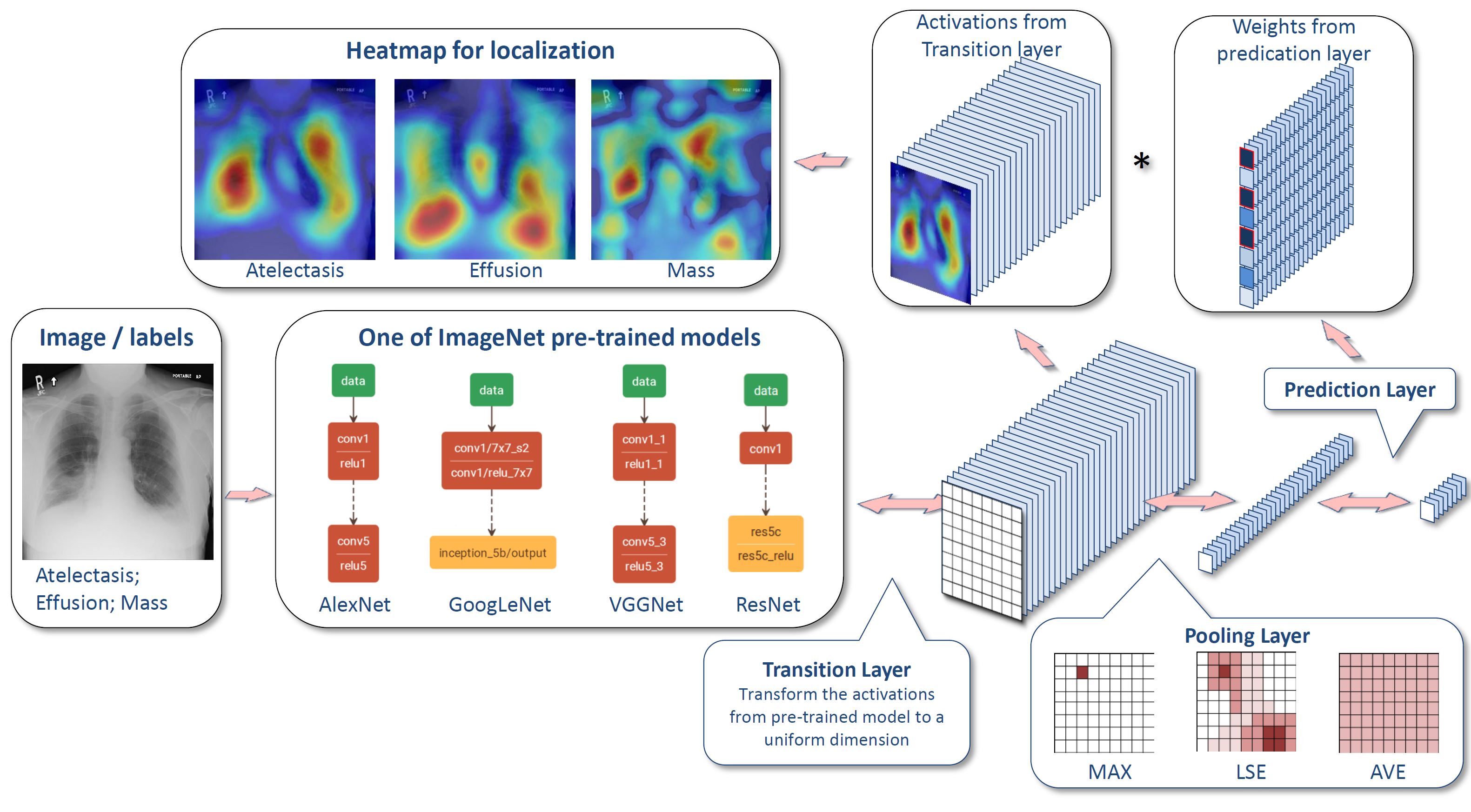}
	\caption{The overall flow-chart of our unified DCNN framework and disease localization process.}
	\label{fig:CNN_Arch}
\end{figure*}

Our goal is to first detect if one or multiple pathologies are presented in each X-ray image and later we can locate them using the activation and weights extracted from the network. We tackle this problem by training a multi-label DCNN classification model. Fig.~\ref{fig:CNN_Arch} illustrates the DCNN architecture we adapted, with similarity to several previous weakly-supervised object localization methods \cite{oquab2015object,zhou2015learning,durandweldon,Hwang2016Self}. As shown in Fig.~\ref{fig:CNN_Arch}, we perform the network surgery on the pre-trained models (using ImageNet~\cite{deng2009imagenet,russakovsky2015imagenet}), e.g., AlexNet~\cite{krizhevsky2012imagenet}, GoogLeNet~\cite{szegedy2015going}, VGGNet-16~\cite{simonyan2014very} and ResNet-50~\cite{He2015}, by leaving out the fully-connected layers and the final classification layers. Instead we insert a transition layer, a global pooling layer, a prediction layer and a loss layer in the end (after the last convolutional layer). In a similar fashion as described in \cite{zhou2015learning}, a combination of deep activations from transition layer (a set of spatial image features) and the weights of prediction inner-product layer (trained feature weighting) can enable us to find the plausible spatial locations of diseases.

{\bf Multi-label Setup:}
 There are several options of image-label representation and the choices of multi-label classification loss functions. Here, we define a 8-dimensional label vector $\textbf{y} = [y_{1},...,y_{c},...,y_{C}], y_{c}\in\{0,1\}, C=8$ for each image. $y_{c}$ indicates the presence with respect to according pathology in the image while a all-zero vector $[0,0,0,0,0,0,0,0]$ represents the status of ``Normal'' (no pathology is found in the scope of any of 8 disease categories as listed). This definition transits the multi-label classification problem into a regression-like loss setting. 

{\bf Transition Layer:}
Due to the large variety of pre-trained DCNN architectures we adopt, a transition layer is usually required to transform the activations from previous layers into a uniform dimension of output, $S\times S \times D, S\in\{8,16,32\}$. $D$ represents the dimension of features at spatial location $(i,j), i,j \in\{1,...,S\}$, which can be varied in different model settings, e.g., $D=1024$ for GoogLeNet and $D=2048$ for ResNet. The transition layer helps pass down the weights from pre-trained DCNN models in a standard form, which is critical for using this layers' activations to further generate the heatmap in pathology localization step. 

{\bf Multi-label Classification Loss Layer:}
We first experiment 3 standard loss functions for the regression task instead of using the softmax loss for traditional multi-class classification model, i.e., Hinge Loss (HL), Euclidean Loss (EL) and Cross Entropy Loss (CEL). However, we find that the model has difficulty learning positive instances (images with pathologies) and the image labels are rather sparse, meaning there are extensively more `0's than `1's. This is due to our one-hot-like image labeling strategy and the unbalanced numbers of pathology and ``Normal'' classes. Therefore, we introduce the positive/negative balancing factor $\beta_{P},\beta_{N}$ to enforce the learning of positive examples. For example, the weighted CEL (W-CEL) is defined as follows,
\begin{equation}
\begin{split}
&L_{W\textendash CEL} (f(\vec{x}),\vec{y})=\\
&\beta_{P}\sum_{y_{c}=1}-\ln(f(x_{c}))+ \beta_{N}\sum_{y_{c}=0}-\ln(1-f(x_{c})),  
\end{split} \label{eq:balanced}
\end{equation}
where $\beta_{P}$ is set to $\frac{|P|+|N|}{|P|}$ while $\beta_{N}$ is set to $\frac{|P|+|N|}{|N|}$. $|P|$ and $|N|$ are the total number of `1's and `0's in a batch of image labels.

\subsection{Weakly-Supervised Pathology Localization}
{\bf Global Pooling Layer and Prediction Layer:}
In our multi-label image classification network, the global pooling and the predication layer are designed not only to be part of the DCNN for classification but also to generate the likelihood map of pathologies, namely a heatmap. The location with a peak in the heatmap generally corresponds to the presence of disease pattern with a high probability. The upper part of Fig.~\ref{fig:CNN_Arch} demonstrates the process of producing this heatmap. By performing a global pooling after the transition layer, the weights learned in the prediction layer can function as the weights of spatial maps from the transition layer. Therefore, we can produce weighted spatial activation maps for each disease class (with a size of $S\times S \times C$) by multiplying the activation from transition layer (with a size of $S\times S \times D$) and the weights of prediction layer (with a size of $D\times C$).

The pooling layer plays an important role that chooses what information to be passed down. Besides the conventional max pooling and average pooling, we also utilize the Log-Sum-Exp (LSE) pooling proposed in \cite{pinheiro2015image}. The LSE pooled value $x_{p}$ is defined as 
\begin{equation}
x_{p} = \frac{1}{r} \cdot \log \left[\frac{1}{S}\cdot\sum_{(i,j)\in \textbf{S}}exp(r\cdot x_{ij})\right],
\label{eq:LSE}
\end{equation}
where $x_{ij}$ is the activation value at $(i,j)$, $(i,j)$ is one location in the pooling region $\textbf{S}$, and $S = s\times s$ is the total number of locations in $\textbf{S}$. By controlling the hyper-parameter $r$, the pooled value ranges from the maximum in $\textbf{S}$ (when $r \rightarrow \infty$) to average ($r \rightarrow 0$). It serves as an adjustable option between max pooling and average pooling. Since the LSE function suffers from overflow/underflow problems, the following equivalent is used while implementing the LSE pooling layer in our own DCNN architecture, 
\begin{equation}
x_{p} = x^{*} + \frac{1}{r} \cdot \log \left[\frac{1}{S}\cdot\sum_{(i,j)\in \textbf{S}}exp(r\cdot (x_{ij}-x^{*})\right],
\label{eq:LSE-imp}
\end{equation}
where $x^{*}=max\{|x_{ij}|,(i,j)\in \textbf{S}\}$.

{\bf Bounding Box Generation:}
The heatmap produced from our multi-label classification framework indicates the approximate spatial location of one particular thoracic disease class each time. Due to the simplicity of intensity distributions in these resulting heatmaps, applying an ad-hoc thresholding based B-Box generation method for this task is found to be sufficient. The intensities in heatmaps are first normalized to $[0,255]$ and then thresholded by $\{60, 180\}$ individually. Finally, B-Boxes are generated to cover the isolated regions in the resulting binary maps. 

\section{Experiments}

{\bf Data:} We evaluate and validate the unified disease classification and localization framework using the proposed ChestX-ray8 database. In total, 108,948 frontal-view X-ray images are in the database, of which 24,636 images contain one or more pathologies. The remaining 84,312 images are normal cases. For the pathology classification and localization task, we randomly shuffled the entire dataset into three subgroups for CNN fine-tuning via Stochastic Gradient Descent (SGD): i.e. training (70\%), validation (10\%) and testing (20\%). We only report the 8 thoracic disease recognition performance on the testing set in our experiments. Furthermore, for the 983 images with 1,600 annotated B-Boxes of pathologies, these boxes are only used as the ground truth to evaluate the disease localization accuracy in testing (not for training purpose).  

\begin{table*}[t]
	\small
	\centering
	\begin{tabular}{|p{6.5em}||c|c|c|c|c|c|c|c|}
		\hline  
		{\bf Setting}   & {\bf Atelectasis} & {\bf Cardiomegaly} & {\bf Effusion} & {\bf Infiltration} & {\bf Mass} & {\bf Nodule} & {\bf Pneumonia} & {\bf Pneumothorax} \\
		\hline\hline
		\multicolumn{9}{|c|}{Initialization with different pre-trained models}\\
		\hline
		{\bf AlexNet}   &0.6458&0.6925&0.6642&0.6041&\textbf{0.5644}&0.6487&0.5493&0.7425 \\
		\hline
		{\bf GoogLeNet}    & 0.6307&0.7056&0.6876&0.6088&0.5363&0.5579&0.5990&0.7824\\
		\hline
		{\bf VGGNet-16}   & 0.6281&0.7084&0.6502&0.5896&0.5103&0.6556&0.5100&0.7516 \\
		\hline
		{\bf ResNet-50}   & \textbf{0.7069}&\textbf{0.8141}&\textbf{0.7362}&\textbf{0.6128}&0.5609&\textbf{0.7164}&\textbf{0.6333}&\textbf{0.7891}\\
		\hline
		\multicolumn{9}{|c|}{Different multi-label loss functions}\\
		\hline
		{\bf CEL}   &0.7064&0.7262&0.7351&0.6084&0.5530&0.6545&0.5164&0.7665\\
		\hline
		{\bf W-CEL}   & 0.7069&0.8141&0.7362&0.6128&0.5609&0.7164&0.6333&0.7891\\
		\hline
	\end{tabular}\label{tab:AUC_all}
	\caption{AUCs of ROC curves for multi-label classification in different DCNN model setting.}
\end{table*}

\begin{table*}[t]
	\small
	\centering
	\begin{tabular}{|p{6.5em}||c|c|c|c|c|c|c|c|}
		\hline
		{\bf T(IoBB)}   & {\bf Atelectasis} & {\bf Cardiomegaly} & {\bf Effusion} & {\bf Infiltration} & {\bf Mass} & {\bf Nodule} & {\bf Pneumonia} & {\bf Pneumothorax} \\
		\hline\hline
		\multicolumn{9}{|c|}{T(IoBB) = 0.1}\\
		\hline
		{\bf Acc.}   & 0.7277&0.9931&0.7124&0.7886&0.4352&0.1645&0.7500&0.4591\\
		\hline
		{\bf AFP}    & 0.8323&0.3506&0.7998&0.5589&0.6423&0.6047&0.9055&0.4776\\
		\hline
		\multicolumn{9}{|c|}{\textbf{T(IoBB) = 0.25} (Two times larger on both x and y axis than ground truth B-Boxes)}\\
		\hline
		{\bf Acc.}   & 0.5500&0.9794&0.5424&0.5772&0.2823&0.0506&0.5583&0.3469\\
		\hline
		{\bf AFP}   & 0.9167&0.4553&0.8598&0.6077&0.6707&0.6158&0.9614&0.5000\\
		\hline
		\multicolumn{9}{|c|}{T(IoBB) = 0.5}\\
		\hline
		{\bf Acc.}   & 0.2833&0.8767&0.3333&0.4227&0.1411&0.0126&0.3833&0.1836\\
		\hline
		{\bf AFP}   & 1.0203&0.5630&0.9268&0.6585&0.6941&0.6189&1.0132&0.5285\\
		\hline
		\multicolumn{9}{|c|}{T(IoBB) = 0.75}\\
		\hline		
		{\bf Acc.}   & 0.1666&0.7260&0.2418&0.3252&0.1176&0.0126&0.2583&0.1020\\
		\hline
		{\bf AFP}   & 1.0619&0.6616&0.9603&0.6921&0.7043&0.6199&1.0569&0.5396\\
		\hline
		\multicolumn{9}{|c|}{T(IoBB) = 0.9} \\
		\hline
		{\bf Acc.}   & 0.1333&0.6849&0.2091&0.2520&0.0588&0.0126&0.2416&0.0816\\
		\hline
		{\bf AFP}   & 1.0752&0.7226&0.9797&0.7124&0.7144&0.6199&1.0732&0.5437\\
		\hline
	\end{tabular}\label{tab:IoBB}
	\caption{Pathology localization accuracy and average false positive number for 8 disease classes.}
\end{table*}

{\bf CNN Setting:}
Our multi-label CNN architecture is implemented using Caffe framework~\cite{jia2014caffe}. The ImageNet pre-trained models, i.e., AlexNet~\cite{krizhevsky2012imagenet}, GoogLeNet~\cite{szegedy2015going}, VGGNet-16~\cite{simonyan2014very} and ResNet-50~\cite{He2015} are obtained from the Caffe model zoo
. Our unified DCNN takes the weights from those models and only the transition layers and prediction layers are trained from scratch.

Due to the large image size and the limit of GPU memory, it is necessary to reduce the image \textit{batch\_size} to load the entire model and keep activations in GPU while we increase the \textit{iter\_size} to accumulate the gradients for more iterations. The combination of both may vary in different CNN models but we set $batch\_size\times iter\_size = 80$ as a constant. Furthermore, the total training iterations are customized for different CNN models to prevent over-fitting. More complex models like ResNet-50 actually take less iterations (e.g., 10000 iterations) to reach the convergence. The DCNN models are trained using a Dev-Box linux server with 4 Titan X GPUs.     

\begin{figure}[t]
\centering
\begin{tabular}{cc}
	\includegraphics[width=0.49\linewidth]{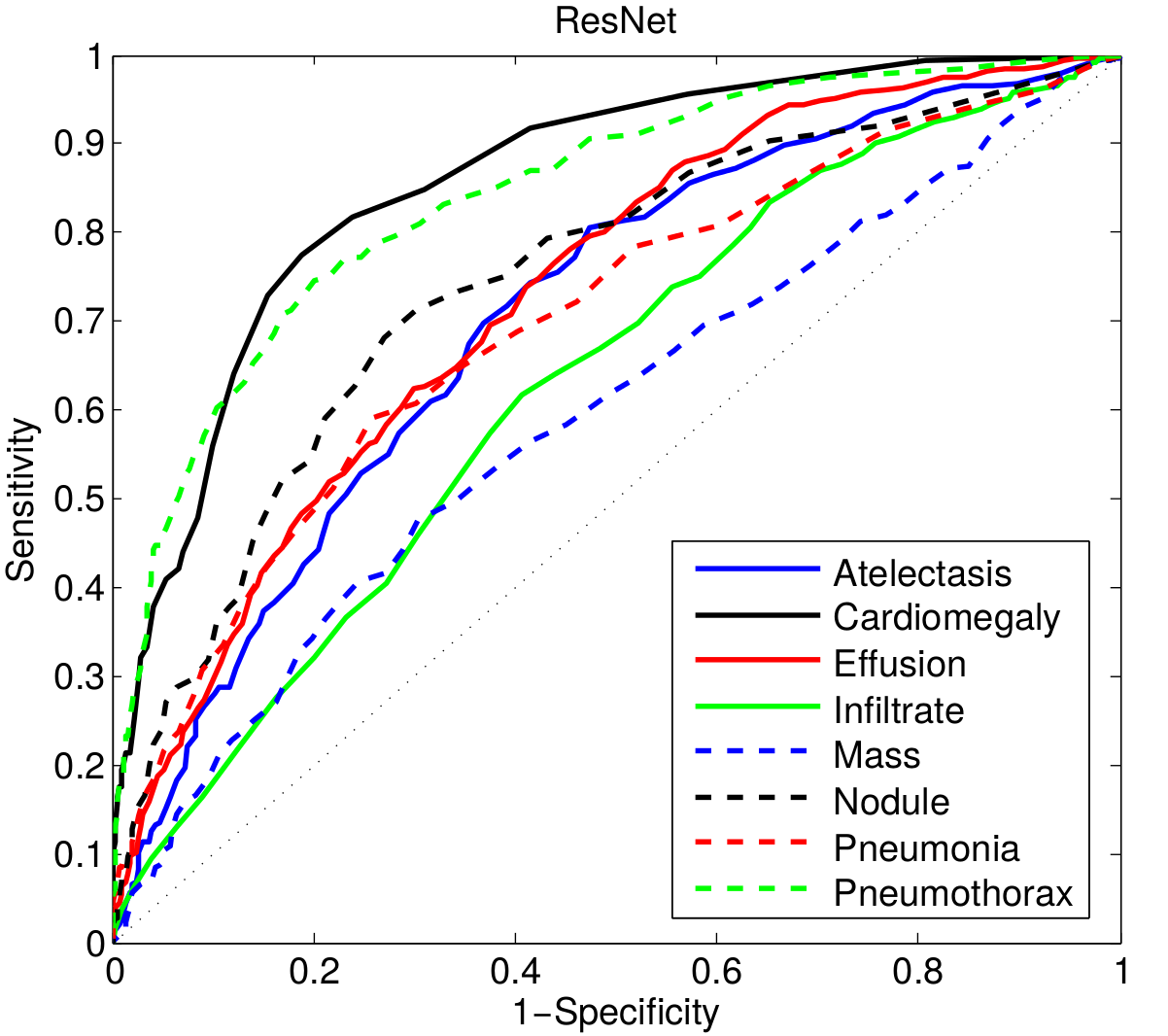} 
	\includegraphics[width=0.49\linewidth]{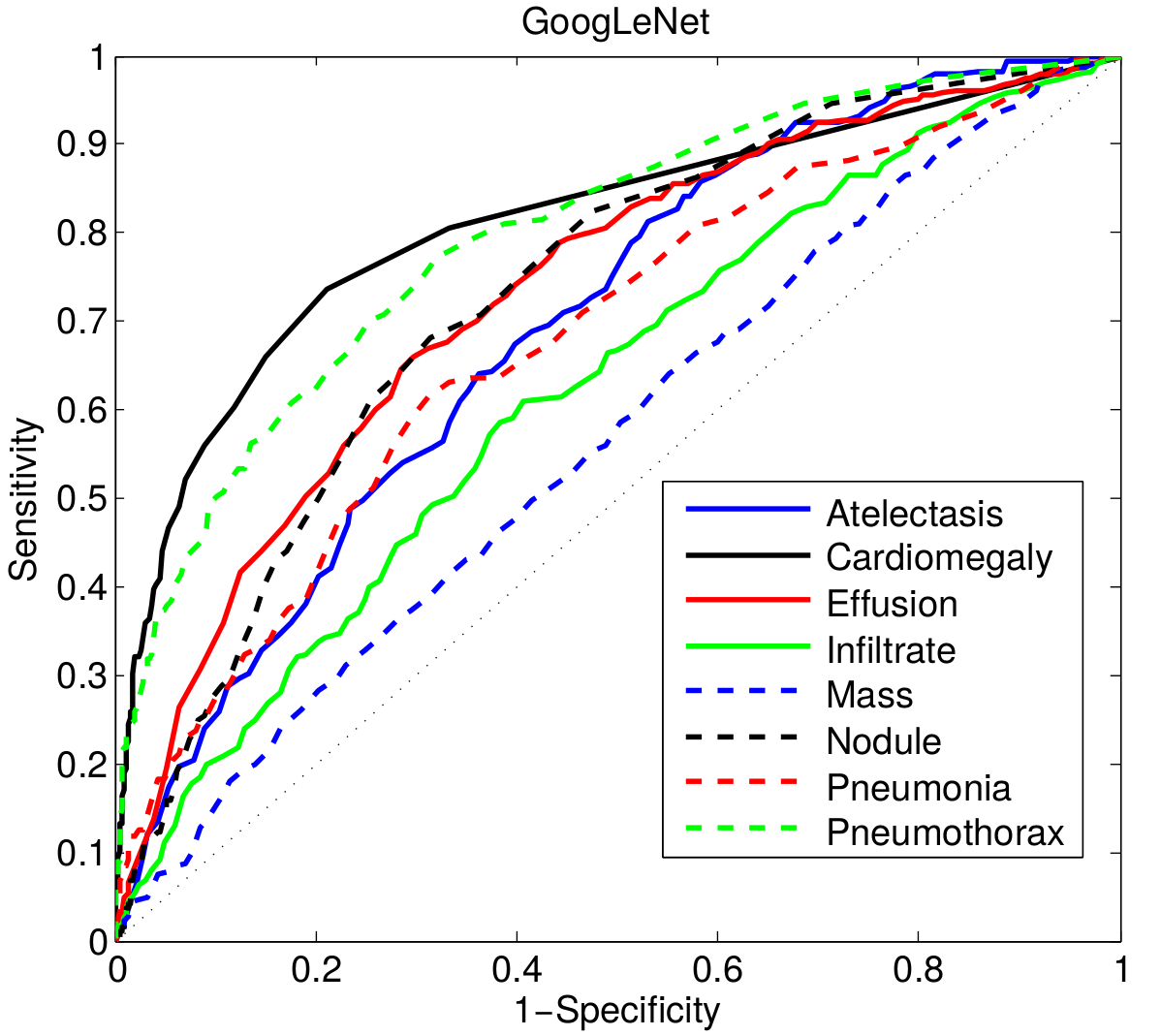} \\
	\includegraphics[width=0.49\linewidth]{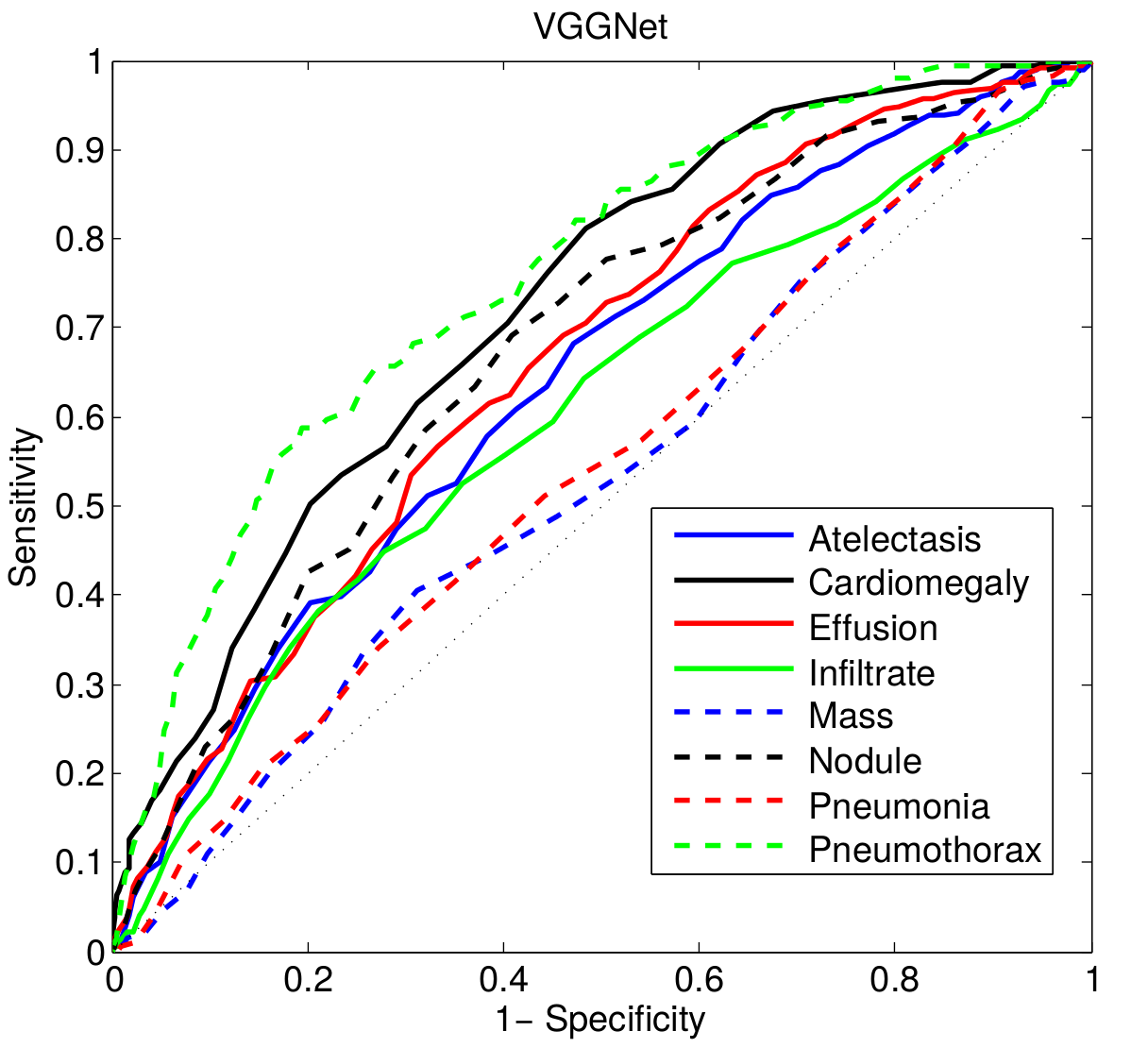} 
	\includegraphics[width=0.49\linewidth]{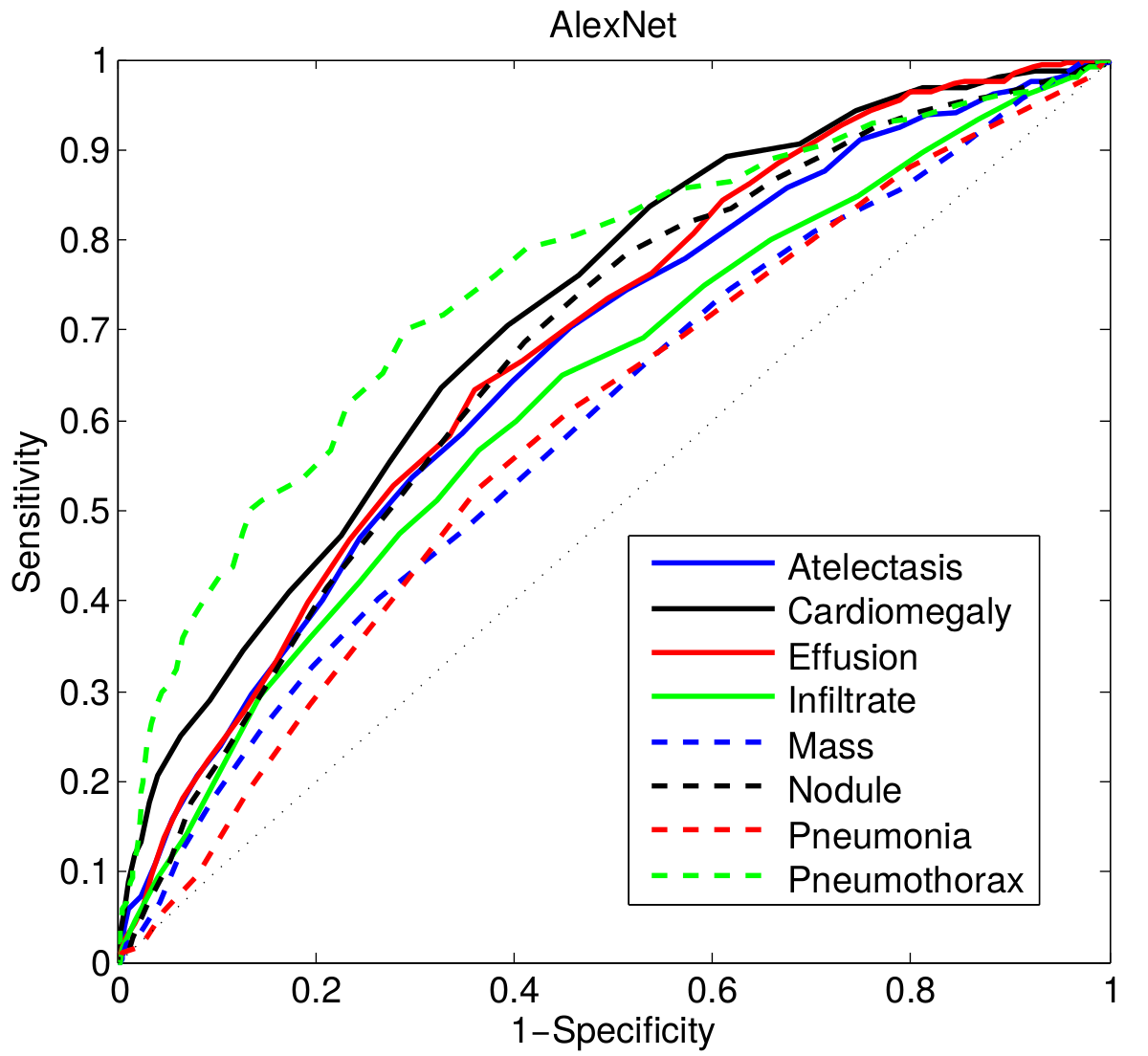} \\
\end{tabular}
	\caption{A comparison of multi-label classification performance with different model initializations.}
	\label{fig:ROC_8_4model}
\end{figure}

{\bf Multi-label Disease Classification:}
Fig.~\ref{fig:ROC_8_4model} demonstrates the multi-label classification ROC curves on 8 pathology classes by initializing the DCNN framework with 4 different pre-trained models of AlexNet, GoogLeNet, VGG and ResNet-50. The corresponding Area-Under-Curve (AUC) values are given in Table~\ref{tab:AUC_all}. The quantitative performance varies greatly, in which the model based on ResNet-50 achieves the best results. The ``Cardiomegaly'' (AUC=0.8141) and ``Pneumothorax'' (AUC=0.7891) classes are consistently well-recognized compared to other groups while the detection ratios can be relatively lower for pathologies which contain small objects, e.g., ``Mass'' (AUC=0.5609) and ``Nodule'' classes. Mass is difficult to detect due to its huge within-class appearance variation. The lower performance on ``Pneumonia'' (AUC=0.6333) is probably because of lack of total instances in our patient population (less than 1\% X-rays labeled as Pneumonia). This finding is consistent with the comparison on object detection performance, degrading from PASCAL VOC \cite{Everingham2015} to MS COCO \cite{Lin2014COCO} where many small annotated objects appear. 

\begin{figure}[h]
\centering
	\includegraphics[width=0.80\columnwidth]{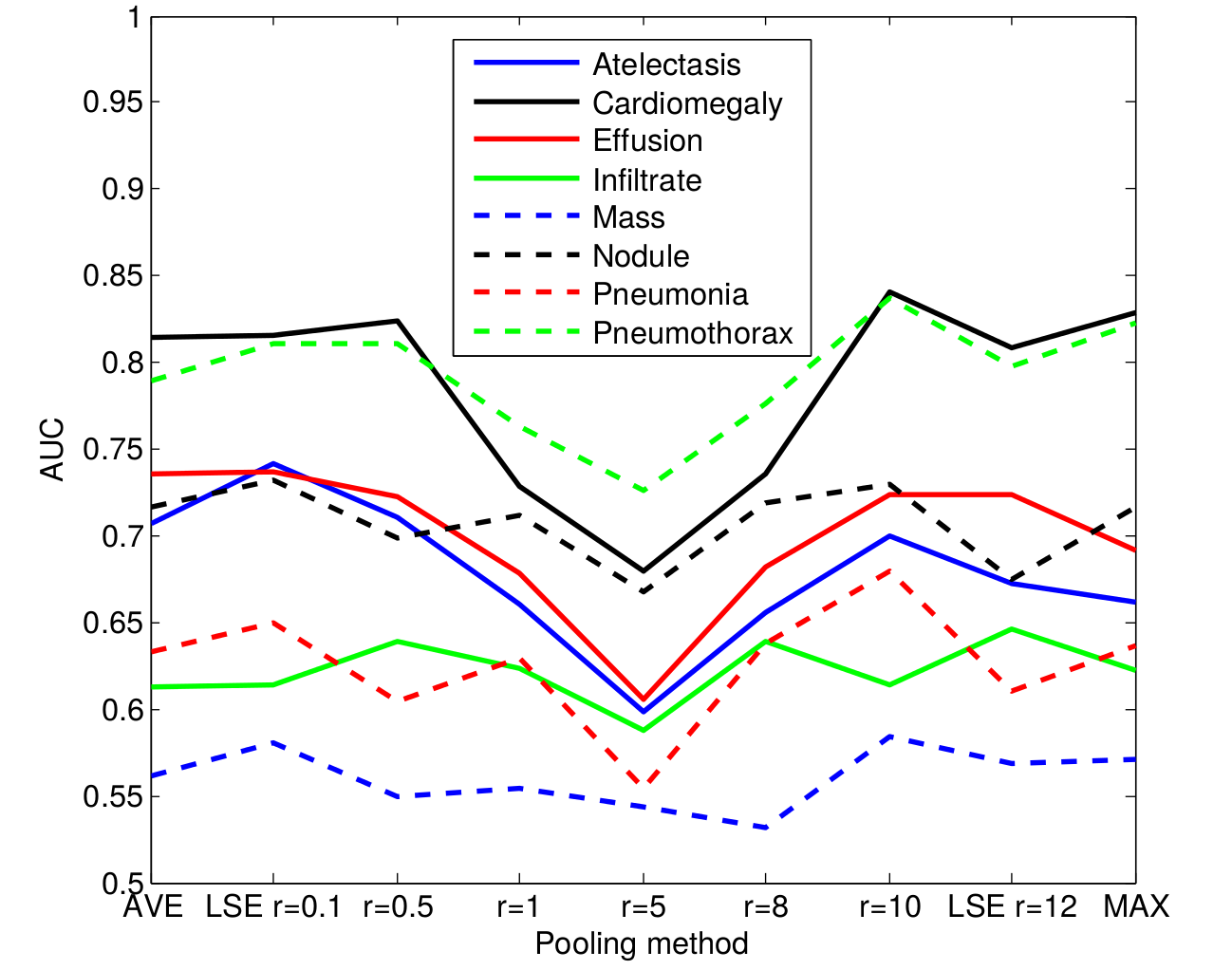}
	\caption{A comparison of multi-label classification performance with different pooling strategies.}
	\label{fig:AUC_pool}
\end{figure}

Next, we examine the influence of different pooling strategies when using ResNet-50 to initialize the DCNN framework. As discussed above, three types of pooling schemes are experimented: average looping, LSE pooling and max pooling. The hyper-parameter $r$ in LSE pooling varies in $\{0.1,0.5,1,5,8,10,12\}$. As illustrated in Fig. \ref{fig:AUC_pool}, average pooling and max pooling achieve approximately equivalent performance in this classification task. The performance of LSE pooling start declining first when $r$ starts increasing and reach the bottom when $r=5$. Then it reaches the overall best performance around $r=10$. LSE pooling behaves like a weighed pooling method or a transition scheme between average and max pooling under different $r$ values. Overall, LSE pooling ($r=10$) reports the best performance (consistently higher than mean and max pooling). 

Last, we demonstrate the performance improvement by using the positive/negative instances balanced loss functions (Eq. \ref{eq:balanced}). As shown in Table \ref{tab:AUC_all}, the weighted loss (W-CEL) provides better overall performance than CEL, especially for those classes with relative fewer positive instances, e.g. AUC for ``Cardiomegaly'' is increased from 0.7262 to 0.8141 and from 0.5164 to 0.6333 for ``Pneumonia''.


{\bf Disease Localization:}
Leveraging the fine-tuned DCNN models for multi-label disease classification, we can calculate the disease heatmaps using the activations of the transition layer and the weights from the prediction layer, and even generate the B-Boxes for each pathology candidate. The computed bounding boxes are evaluated against the hand annotated ground truth (GT) boxes (included in ChestX-ray8). Although the total number of B-Box annotations (1,600 instances) is relatively small compared to the entire dataset, it may be still sufficient to get a reasonable estimate on how the proposed framework performs on the weakly-supervised disease localization task. To examine the accuracy of computerized B-Boxes versus the GT B-Boxes, two types of measurement are used, i.e, the standard Intersection over Union ratio (IoU) or the Intersection over the detected B-Box area ratio (IoBB) (similar to Area of Precision or Purity). Due to the relatively low spatial resolution of heatmaps ($32\times 32$) in contrast to the original image dimensions ($1024\times 1024$), the computed B-Boxes are often larger than the according GT B-Boxes. Therefore, we define a correct localization by requiring either $IoU > T(IoU)$ or $IoBB > T(IoBB)$. Refer to the supplementary material for localization performance under varying $T(IoU)$. Table~\ref{tab:IoBB} illustrates the localization accuracy (Acc.) and Average False Positive (AFP) number for each disease type, with $T(IoBB)\in \{0.1, 0.25, 0.5, 0.75, 0.9\}$. Please refer to the supplementary material for qualitative exemplary disease localization results for each of 8 pathology classes. 


\section{Conclusion}
Constructing hospital-scale radiology image databases with computerized diagnostic performance benchmarks has not been addressed until this work. We attempt to build a ``machine-human annotated'' comprehensive chest X-ray database that presents the realistic clinical and methodological challenges of handling at least tens of thousands of patients (somewhat similar to ``ImageNet'' in natural images). We also conduct extensive quantitative performance benchmarking on eight common thoracic pathology classification and weakly-supervised localization using ChestX-ray8 database. The main goal is to initiate future efforts by promoting public datasets in this important domain. Building truly large-scale, fully-automated high precision medical diagnosis systems remains a strenuous task. ChestX-ray8 can enable the data-hungry deep neural network paradigms to create clinically meaningful applications, including common disease pattern mining, disease correlation analysis, automated radiological report generation, etc. For future work, ChestX-ray8 will be extended to cover more disease classes and integrated with other clinical information, e.g., follow-up studies across time and patient history.

{\bf Acknowledgements  } This work was supported by the Intramural Research Programs of the NIH Clinical Center and National Library of Medicine. We thank NVIDIA Corporation for the GPU donation.

{\small
\bibliographystyle{ieee}
\bibliography{egbib_wxs}
}
\vspace{130mm}

\appendix

\section{Supplementary Materials}
\subsection{SNOMED-CT Concepts}
In this work, we only consider the semantic types of Diseases or Syndromes and Findings (namely `dsyn' and `fndg' respectively). Table~\ref{tab:cuis} shows the corresponding SNOMED-CT concepts that are relevant to the target diseases (these mappings are developed by searching the disease names in the~\umls terminology service~\footnote{\url{https://uts.nlm.nih.gov/metathesaurus.html}}, and verified by a board-certificated radiologist.

\begin{table}[b]
	\begin{center}
		\begin{tabular}{ll@{\hspace{4em}}l}
			\hline
			\multicolumn{2}{l}{CUI} & Concept\\\hline\hline
			\multicolumn{3}{l}{Atelectasis} \\
			& C0004144 & atelectasis\\
			& C0264494 & discoid atelectasis\\
			& C0264496 & focal atelectasis\\
			\multicolumn{3}{l}{Cardiomegaly}\\
			& C0018800 & cardiomegaly\\
			\multicolumn{3}{l}{Effusion}\\
			& C0013687 & effusion\\
			& C0031039 & pericardial effusion\\
			& C0032227 & pleural effusion disorder\\
			& C0747635 & bilateral pleural effusion\\
			& C0747639 & loculated pleural effusion\\
			\multicolumn{3}{l}{Pneumonia}\\
			& C0032285 & pneumonia\\
			& C0577702 & basal pneumonia\\
			& C0578576 & left upper zone pneumonia\\
			& C0578577 & right middle zone pneumonia\\
			& C0585104 & left lower zone pneumonia\\
			& C0585105 & right lower zone pneumonia\\
			& C0585106 & right upper zone pneumonia\\
			& C0747651 & recurrent aspiration pneumonia\\
			& C1960024 & lingular pneumonia\\
			\multicolumn{3}{l}{Pneumothorax}\\
			& C0032326 & pneumothorax\\
			& C0264557 & chronic pneumothorax\\
			& C0546333 & right pneumothorax\\
			& C0546334 & left pneumothorax\\
			\hline
		\end{tabular}
	\end{center}
	\caption{Sample Target Diseases and their corresponding concept and identifiers (CUIs) in SNOMED-CT.}
	\label{tab:cuis}
\end{table}

\subsection{Rules of Negation/Uncertainty}
Although many text processing systems can handle the negation/uncertainty detection problem, most of them exploit regular expressions on the text directly. One of the disadvantages to use regular expressions for negation/uncertainty detection is that they cannot capture various syntactic constructions for multiple subjects. For example, in the phrase of ``clear of A and B'', the regular expression can capture ``A'' as a negation but not ``B'', particularly when both ``A'' and ``B'' are long and complex noun phrases. 

To overcome this complication, we hand-craft a number of novel rules of negation/uncertainty defined on the syntactic level in this work. More specifically, we utilize the syntactic dependency information because it is close to the semantic relationship between words and thus has become prevalent in biomedical text processing. We defined our rules on the dependency graph, by utilizing the dependency label and direction information between words. Table~\ref{tab:rules} shows the rules we  defined  for negation/uncertainty detection on the syntactic level.

\begin{table*}[t]
	\begin{center}
		\begin{tabular}{ll|l}
			\hline
			\multicolumn{2}{l|}{Rule} & Example\\
			\hline\hline
			\multicolumn{3}{l}{Negation} \\
			\hline
			& no$\leftarrow * \leftarrow$ DISEASE & No acute pulmonary disease\\
			
			& $* \rightarrow$ \textit{prep\_without} $\rightarrow$ DISEASE & changes without focal airspace disease\\
			
			& clear/free/disappearance $\rightarrow$ \textit{prep\_of} $\rightarrow$ DISEASE & clear of focal airspace disease, pneumothorax, or pleural effusion\\
			
			& $* \rightarrow$ \textit{prep\_without} $\rightarrow$ evidence $\rightarrow$ \textit{prep\_of} $\rightarrow$ DISEASE & Changes without evidence of acute infiltrate \\
			
			& no $\leftarrow$ \textit{neg} $\leftarrow$ evidence $\rightarrow$ \textit{prep\_of} $\rightarrow$ DISEASE & No evidence of active disease \\
			\hline
			\multicolumn{3}{l}{Uncertainty}\\
			\hline
			& cannot $\leftarrow$ \textit{md} $\leftarrow$ exclude & The aorta is tortuous, and cannot exclude ascending aortic aneurysm\\
			& concern $\rightarrow$ \textit{prep\_for} $\rightarrow *$ & There is raises concern for pneumonia \\
			& could be/may be/... & which could be due to nodule/lymph node \\
			& difficult $\rightarrow$ \textit{prep\_to} $\rightarrow$ exclude & interstitial infiltrates difficult to exclude \\
			& may $\leftarrow$ \textit{md} $\leftarrow$ represent & which may represent pleural reaction or small pulmonary nodules \\
			& suggesting/suspect/... $\rightarrow$ \textit{dobj} $\rightarrow$ DISEASE & Bilateral pulmonary nodules suggesting pulmonary metastases \\
			\hline
		\end{tabular}
	\end{center}
	\caption{Rules and corresponding examples for negation and uncertainty detection.}
	\label{tab:rules}
\end{table*}

\subsection{More Disease Localization Results}
Table~\ref{tab:IoU} illustrates the localization accuracy (Acc.) and Average False Positive (AFP) number for each disease type, with  $IoU > T(IoU)$ only and $T(IoU)\in \{0.1, 0.2, 0.3, 0.4, 0.5, 0.6,0.7\}$. 

\begin{table*}[t]
	
	\begin{center}
		\begin{tabular}{|p{6.5em}||c|c|c|c|c|c|c|c|}
			\hline
			{\bf T(IoU)}   & {\bf Atelectasis} & {\bf Cardiomegaly} & {\bf Effusion} & {\bf Infiltration} & {\bf Mass} & {\bf Nodule} & {\bf Pneumonia} & {\bf Pneumothorax} \\
			\hline\hline
			\multicolumn{9}{|c|}{T(IoU) = 0.1}\\
			\hline
			{\bf Acc.}   & 0.6888&0.9383&0.6601&0.7073&0.4000&0.1392&0.6333&0.3775\\
			\hline
			{\bf AFP}    & 0.8943&0.5996&0.8343&0.6250&0.6666&0.6077&1.0203&0.4949\\
			\hline
			\multicolumn{9}{|c|}{T(IoU) = 0.2 }\\
			\hline
			{\bf Acc.}   & 0.4722&0.6849&0.4509&0.4796&0.2588&0.0506&0.3500&0.2346\\
			\hline
			{\bf AFP}   & 0.9827&0.7205&0.9096&0.6849&0.6941&0.6158&1.0793&0.5173\\
			\hline
			\multicolumn{9}{|c|}{T(IoU) = 0.3}\\
			\hline
			{\bf Acc.}   &0.2444&0.4589&0.3006&0.2764&0.1529&0.0379&0.1666&0.1326\\
			\hline
			{\bf AFP}   & 1.0417&0.7815&0.9472&0.7236&0.7073&0.6168&1.1067&0.5325\\
			\hline
			\multicolumn{9}{|c|}{T(IoU) = 0.4}\\
			\hline
			{\bf Acc.}   &0.0944&0.2808&0.2026&0.1219&0.0705&0.0126&0.0750&0.0714\\
			\hline
			{\bf AFP}   & 1.0783&0.8140&0.9705&0.7489&0.7164&0.6189&1.1239&0.5427\\
			\hline
			\multicolumn{9}{|c|}{T(IoU) = 0.5 }\\
			\hline
			{\bf Acc.}   & 0.0500&0.1780&0.1111&0.0650&0.0117&0.0126&0.0333&0.0306\\
			\hline
			{\bf AFP}   & 1.0884&0.8354&0.9919&0.7571&0.7215&0.6189&1.1291&0.5478\\
			\hline
			\multicolumn{9}{|c|}{T(IoU) = 0.6}\\
			\hline		
			{\bf Acc.}   &0.0222&0.0753&0.0457&0.0243&0.0000&0.0126&0.0166&0.0306\\
			\hline
			{\bf AFP}   & 1.0935&0.8506&1.0051&0.7632&0.7226&0.6189&1.1321&0.5478\\
			\hline
			\multicolumn{9}{|c|}{T(IoU) = 0.7} \\
			\hline
			{\bf Acc.}   &0.0055&0.0273&0.0196&0.0000&0.0000&0.0000&0.0083&0.0204\\
			\hline
			{\bf AFP}   & 1.0965&0.8577&1.009&0.7663&0.7226&0.6199&1.1331&0.5488\\
			\hline
		\end{tabular}
	\end{center}
	\caption{Pathology localization accuracy and average false positive number for 8 disease classes with $T(IoU)$ ranged from $0.1$ to $0.7$.}
	\label{tab:IoU}
\end{table*}

Table \ref{tab:Loc_example_1} to Table \ref{tab:Loc_example_8} illustrate localization results from each of 8 disease classes together with associated report and mined disease keywords. The heatmaps overlay on the original images are shown on the right. Correct bounding boxes (in green), false positives (in red) and the groundtruth (in blue) are plotted over the original image on the left.

In order to quantitatively demonstrate how informative those heatmaps are, a simple two-level thresholding based bounding box generator is adopted here to catch the peaks in the heatmap and later generated bounding boxes can be evaluated against the ground truth. Each heatmap will approximately results in 1-3 bounding boxes. We believe the localization accuracy and AFP (shown in Table \ref{tab:IoU}) could be further optimized by adopting a more sophisticated bounding box generation method, e.g. selective search \cite{uijlings2013selective} or Edgebox \cite{hosang2016makes}. Nevertheless, we reserve the effort to do so, since our main goal is not to compute the exact spatial location of disease patterns but just to obtain some instructive location information for future applications, e.g. automated radiological report generation. Take the case shown in Table \ref{tab:Loc_example_1} for an example. The peak at the lower part of the left lung region indicates the presence of ``atelectasis", which confer the statement of ``...stable abnormal study including left basilar infilrate/atelectasis, ..." presented in the impression section of the associated radiological report. By combining with other information, e.g. a lung region mask, the heatmap itself is already more informative than just the presence indication of certain disease in an image as introduced in the previous works, e.g.~\cite{Shin2016Learning}.   

\begin{table*}
	\begin{center}
		\begin{tabular}{p{15em}|p{6em}|p{23em}}
			\hline
			Radiology report & Keyword & Localization Result\\
			\hline\hline
			findings include: 1. left basilar atelectasis/consolidation. 2. prominent hilum (mediastinal adenopathy). 3. left pic catheter (tip in atriocaval junction). 4. stable, normal appearing cardiomediastinal silhouette. 
			
			impression: small right pleural effusion otherwise stable abnormal study including left basilar infiltrate/atelectasis, prominent hilum, and position of left pic catheter (tip atriocaval junction).
			& Effusion;
			
			Infiltration;
			
			Atelectasis 
			&\vspace{0cm}\includegraphics[width=1\linewidth]{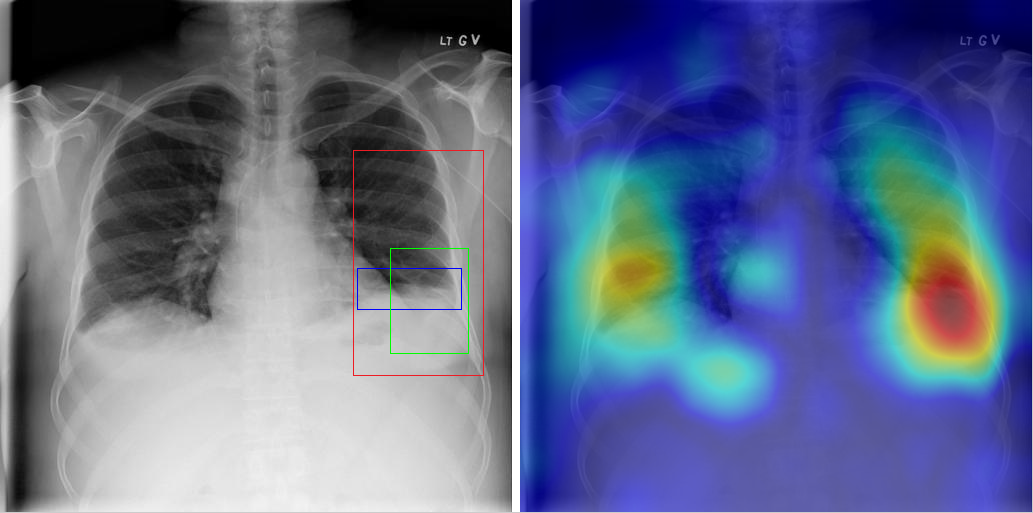} \\
			\hline
		\end{tabular}
	\end{center}
	\caption{A sample of chest x-ray radiology report, mined disease keywords and localization result from the ``Atelectasis" Class. Correct bounding box (in green), false positives (in red) and the ground truth (in blue) are plotted over the original image.}
	\label{tab:Loc_example_1}
\end{table*}

\begin{table*}
	\begin{center}
		\begin{tabular}{p{15em}|p{6em}|p{23em}}
			\hline
			Radiology report & Keyword & Localization Result\\
			\hline\hline
			findings include:
			1. cardiomegaly (ct ratio of 17/30).
			2. otherwise normal lungs and mediastinal contours.
			3. no evidence of focal bone lesion.
			dictating 
			& Cardiomegaly 
			&\vspace{0cm}\includegraphics[width=1\linewidth]{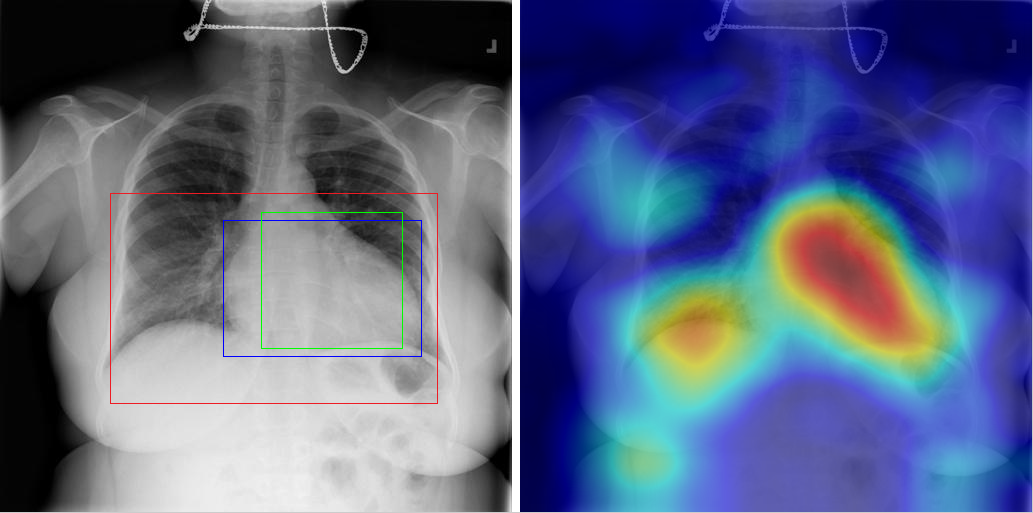} \\
			\hline
		\end{tabular}
	\end{center}
	\caption{A sample of chest x-ray radiology report, mined disease keywords and localization result from the ``Cardiomegaly" Class. Correct bounding box (in green), false positives (in red) and the ground truth (in blue) are plotted over the original image.}
	\label{tab:Loc_example_2}
\end{table*}

\begin{table*}
	\begin{center}
		\begin{tabular}{p{15em}|p{6em}|p{23em}}
			\hline
			Radiology report & Keyword & Localization Result\\
			\hline\hline
			findings: no appreciable change since XX/XX/XX. small right pleural effusion. elevation right hemidiaphragm. diffuse small nodules throughout the lungs, most numerous in the left mid and lower lung. impression: no change with bilateral small lung metastases. 
			& Effusion;
			
			Nodule
			&\vspace{0cm}\includegraphics[width=1\linewidth]{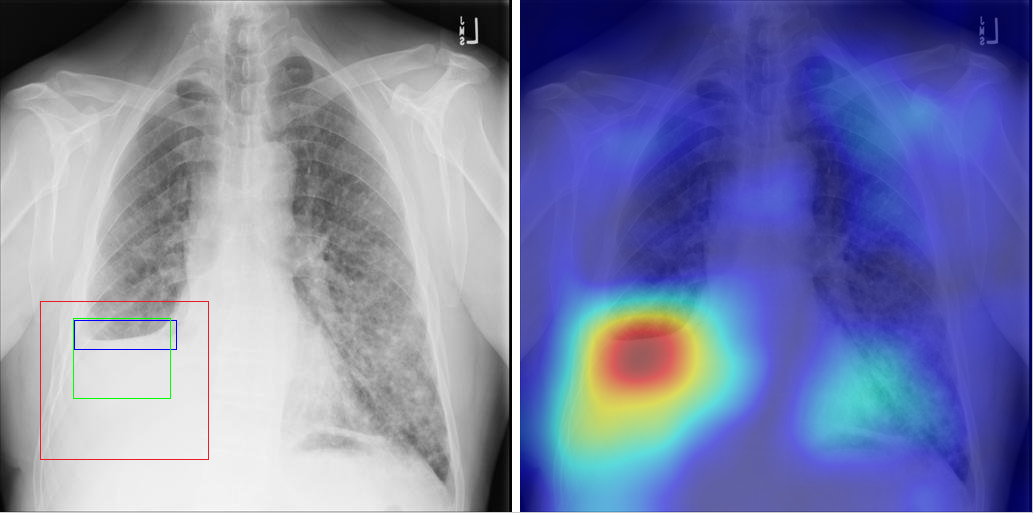} \\
			\hline
		\end{tabular}
	\end{center}
	\caption{A sample of chest x-ray radiology report, mined disease keywords and localization result from the ``Effusion" Class. Correct bounding box (in green), false positives (in red) and the ground truth (in blue) are plotted over the original image.}
	\label{tab:Loc_example_3}
\end{table*}

\begin{table*}
	\begin{center}
		\begin{tabular}{p{15em}|p{6em}|p{23em}}
			\hline
			Radiology report & Keyword & Localization Result\\
			\hline\hline
			findings: port-a-cath reservoir remains in place on the right. chest tube remains in place, tip in the left apex. no pneumothorax. diffuse patchy infiltrates bilaterally are decreasing. 
			
			impression: infiltrates and effusions decreasing.
			& Infiltration 
			&\vspace{0cm}\includegraphics[width=1\linewidth]{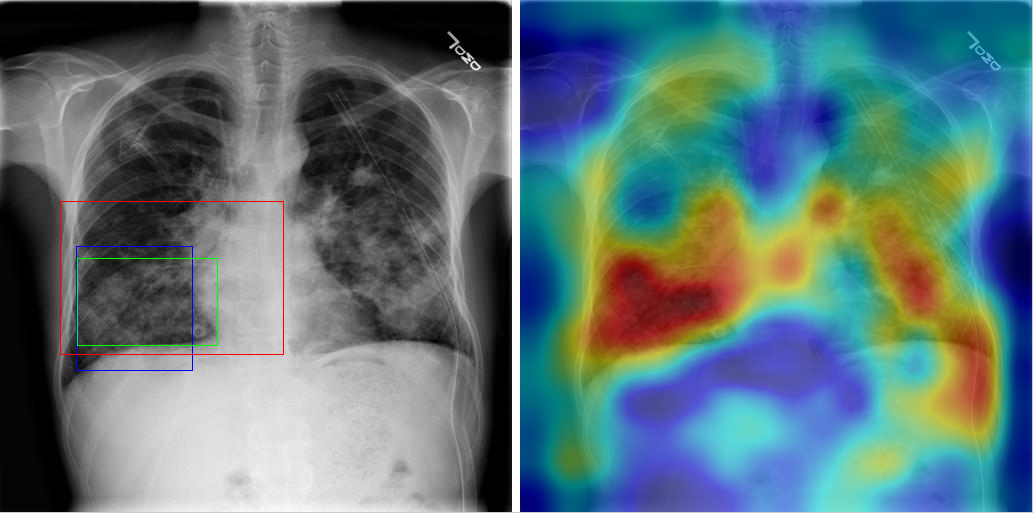} \\
			\hline
		\end{tabular}
	\end{center}
	\caption{A sample of chest x-ray radiology report, mined disease keywords and localization result from the ``Infiltration" Class. Correct bounding box (in green), false positives (in red) and the ground truth (in blue) are plotted over the original image.}
	\label{tab:Loc_example_4}
\end{table*}

\begin{table*}
	\begin{center}
		\begin{tabular}{p{15em}|p{6em}|p{23em}}
			\hline
			Radiology report & Keyword & Localization Result\\
			\hline\hline
			findings: right internal jugular catheter remains in place. large metastatic lung mass in the lateral left upper lobe is again noted. no infiltrate or effusion. extensive surgical clips again noted left axilla. 
			
			impression: no significant change.
			& Mass
			&\vspace{0cm}\includegraphics[width=1\linewidth]{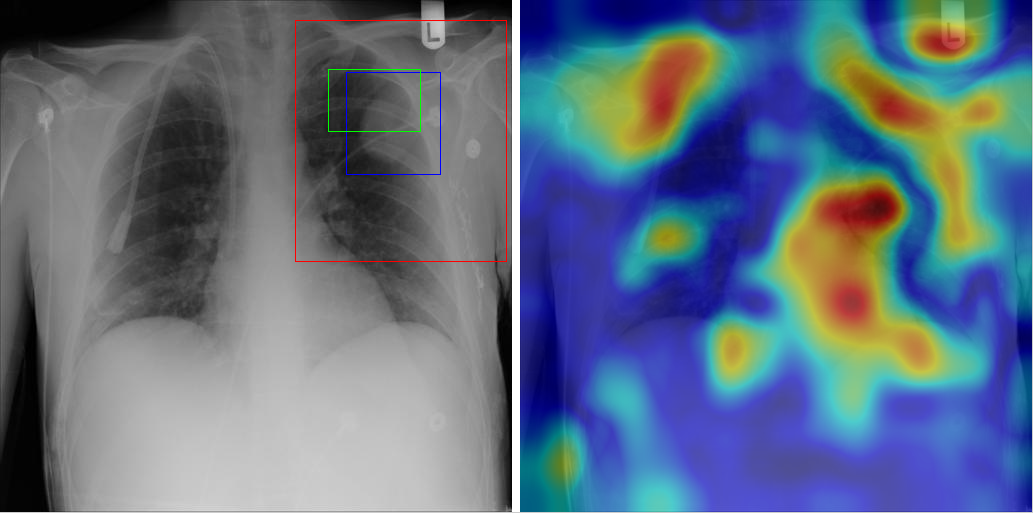} \\
			\hline
		\end{tabular}
	\end{center}
	\caption{A sample of chest x-ray radiology report, mined disease keywords and localization result from the ``Mass" Class. Correct bounding box (in green), false positives (in red) and the ground truth (in blue) are plotted over the original image.}
	\label{tab:Loc_example_5}
\end{table*}

\begin{table*}
	\begin{center}
		\begin{tabular}{p{15em}|p{6em}|p{23em}}
			\hline
			Radiology report & Keyword & Localization Result\\
			\hline\hline
			findings: pa and lateral views of the chest demonstrate stable 2.2 cm nodule in left lower lung field posteriorly. the lungs are clear without infiltrate or effusion. cardiomediastinal silhouette is normal size and contour. pulmonary vascularity is normal in caliber and distribution. 
			
			impression: stable left likely hamartoma.
			& Nodule;
			
			Infiltration
			&\vspace{0cm}\includegraphics[width=1\linewidth]{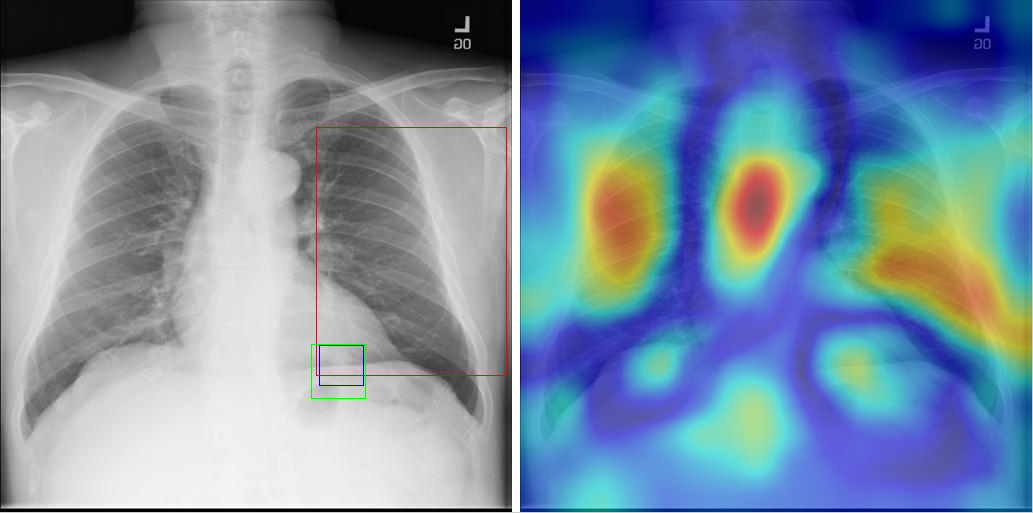} \\
			\hline
		\end{tabular}
	\end{center}
	\caption{A sample of chest x-ray radiology report, mined disease keywords and localization result from the ``Nodule" Class. Correct bounding box (in green), false positives (in red) and the ground truth (in blue) are plotted over the original image.}
	\label{tab:Loc_example_6}
\end{table*}

\begin{table*}
	\begin{center}
		\begin{tabular}{p{15em}|p{6em}|p{23em}}
			\hline
			Radiology report & Keyword & Localization Result\\
			\hline\hline
			findings: unchanged left lower lung field infiltrate/air bronchograms.
			unchanged right perihilar infiltrate with obscuration of the right
			heart border. no evidence of new infiltrate. no evidence of
			pneumothorax the cardiac and mediastinal contours are stable.
			impression:
			1. no evidence pneumothorax.
			2. unchanged left lower lobe and left lingular
			consolidation/bronchiectasis.
			3. unchanged right middle lobe infiltrate 
			& Pneumonia;
			
			Infiltration
			&\vspace{0cm}\includegraphics[width=1\linewidth]{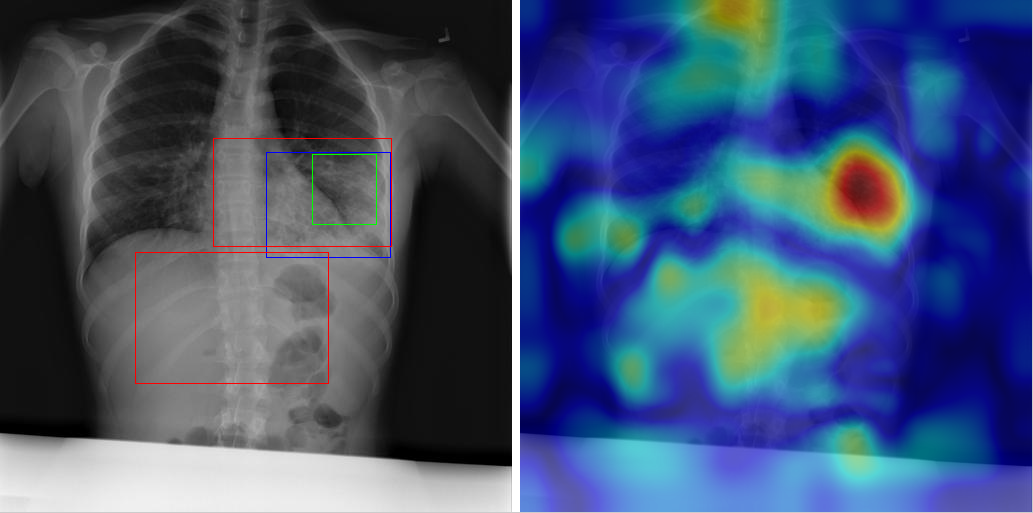} \\
			\hline
		\end{tabular}
	\end{center}
	\caption{A sample of chest x-ray radiology report, mined disease keywords and localization result from the ``Pneumonia" Class. Correct bounding box (in green), false positives (in red) and the ground truth (in blue) are plotted over the original image.}
	\label{tab:Loc_example_7}
\end{table*}

\begin{table*}
	\begin{center}
		\begin{tabular}{p{15em}|p{6em}|p{23em}}
			\hline
			Radiology report & Keyword & Localization Result\\
			\hline\hline
			findings: frontal lateral chest x-ray performed in expiration. left
			apical pneumothorax visible. small pneumothorax visible along the
			left heart border and left hemidiaphragm.
			pleural thickening, mass right chest. the mediastinum cannot be
			evaluated in the expiration. bony structures intact.
			
			impression: left post biopsy pneumothorax.
			& Mass;
			
			Pneumothorax
			&\vspace{0cm}\includegraphics[width=1\linewidth]{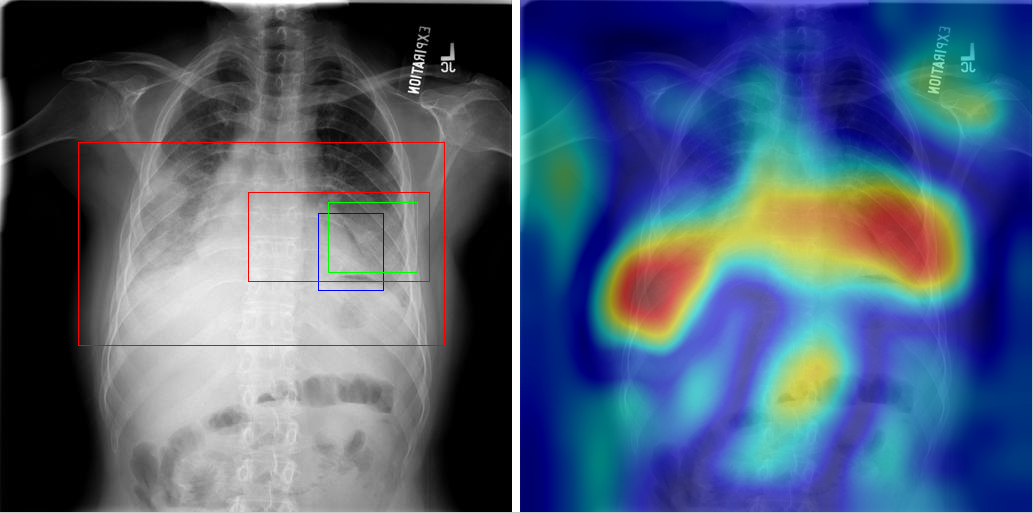} \\
			\hline
		\end{tabular}
	\end{center}
	\caption{A sample of chest x-ray radiology report, mined disease keywords and localization result from the ``Pneumothorax" Class. Correct bounding box (in green), false positives (in red) and the ground truth (in blue) are plotted over the original image.}
	\label{tab:Loc_example_8}
\end{table*}

\cleardoublepage

\hspace{1mm}   
\newpage
\hspace{1mm}
\newpage
\section{ChestX-ray14 Dataset}
After the CVPR submission, we expand the disease categories to include 6 more common thorax diseases (i.e. Consolidation, Edema, Emphysema, Fibrosis, Pleural Thickening and Hernia) and update the NLP mined labels. The statistics of ChestX-ray14 dataset are illustrated in Table \ref{tab:openi_corpus_14} and Figure {\color{red} 8}. The bounding boxes for Pathologies are unchanged at this point.

\begin{table}[h]
	\begin{center}
		\begin{tabular}{l|r|r|r|r|r}
			\hline
			Item \# &X-ray8 & Ov.& X-ray14 & Ov.\\
			\hline\hline
			Report &  108,948 & -& 112,120 &-\\
			\hline
			Atelectasis &  5,789 & 3,286 & 11,535 & 7,323 \\
			Cardiomegaly &  1,010 & 475& 2,772&1,678\\
			Effusion &  6,331 & 4,017& 13,307&9,348\\
			Infiltration &  10,317 & 4,698& 19,871&10,319\\
			Mass &  6,046 & 3,432& 5,746&2,138\\
			Nodule &  1,971 & 1,041 & 6,323&3,617\\
			Pneumonia &  1,062 & 703& 1,353&1,046\\
			Pneumothorax &  2,793 & 1,403& 5,298&3,099\\
			Consolidation &- &- & 4,667&3,353\\
			Edema &- &- & 2,303&1,669\\
			Emphysema &- &- & 2,516&1,621\\
			Fibrosis&- &- & 1,686&959\\
			PT&- &- & 3,385&2,258\\
			Hernia&- &- & 227&117\\
			No findings & 84,312 & 0& 60,412&0 \\
			\hline
		\end{tabular}
	\end{center}
	\caption{Total number (\#) and \# of Overlap (Ov.) of the corpus in ChestX-ray8 and ChestX-ray14 datasets.PT: Pleural Thickening}
	\label{tab:openi_corpus_14}
\end{table}

\subsection{ Evaluation of NLP Mined Labels}
To validate our method, we perform the following experiments. First, we resort to some existing annotated corpora as an alternative, i.e. OpenI dataset. Furthermore, we annotated clinical reports suitable for evaluating finding recognition systems. We randomly selected 900 reports and asked two annotators to mark the above 14 types of findings. Each report was annotated
by two annotators independently and then agreements are reached for conflicts.

Table {\color{red} 18} 
shows the results of our method using OpenI and our proposed dataset, measured in precision (P), recall (R), and F1-score. Much higher precision, recall and F1-scores are achieved compared to the existing MetaMap approach (with NegEx enabled). This indicates that the usage of negation and uncertainty detection on syntactic level successfully removes false positive cases. 

\begin{table}[th]
	\centering
	\begin{tabular}{|p{6.5em}||c|c|}
		\hline
		{\bf ResNet-50}   & ChestX-ray8 & ChestX-ray14 \\
		\hline\hline
		{\bf Atelectasis} & 0.7069& 0.7003 \\
		{\bf Cardiomegaly} & 0.8141& 0.8100 \\
		{\bf Effusion} & 0.7362& 0.7585 \\
		{\bf Infiltration} & 0.6128& 0.6614 \\
		{\bf Mass} & 0.5609& 0.6933 \\
		{\bf Nodule} & 0.7164& 0.6687\\
		{\bf Pneumonia} & 0.6333& 0.6580\\
		{\bf Pneumothorax} & 0.7891 &0.7993\\
		{\bf Consolidation} & - & 0.7032\\
		{\bf Edema} & - & 0.8052\\
		{\bf Emphysema} & - & 0.8330\\
		{\bf Fibrosis} & - & 0.7859\\
		{\bf PT} & - & 0.6835\\
		{\bf Hernia} & - & 0.8717\\
		\hline
	\end{tabular}
	\label{tab:AUC14}
	\caption{AUCs of ROC curves for multi-label classification for ChestX-ray14 using published data split. PT: Pleural Thickening}
\end{table}

\begin{table*}[t]
	\begin{center}
		\begin{tabular}{lr@{~/~}r@{~/~}rr@{~/~}r@{~/~}r}
			\hline
			\multirow{2}{*}{Disease} & \multicolumn{3}{c}{MetaMap} & \multicolumn{3}{c}{Our Method}\\\cline{2-7}
			& Precision & Recall & F1-score & Precision & Recall & F1-score\\\hline\hline
			\multicolumn{7}{c}{OpenI}\\
			\hline
			Atelectasis & 87.3 &96.5 &91.7 & 88.7 & 96.5 & 92.4\\
			Cardiomegaly & 100.0 &85.5 &92.2 & 100.0 &85.5 &92.2\\
			Effusion & 90.3 &87.5 &88.9  & 96.6 &87.5 &91.8\\
			Infiltration & 68.0 &100.0 &81.0 & 81.0 &100.0 &89.5\\
			Mass & 100.0 &66.7 &80.0 & 100.0 &66.7 &80.0\\
			Nodule & 86.7 &65.0 &74.3  & 82.4 &70.0 &75.7\\
			Pneumonia & 40.0 &80.0 &53.3 & 44.4 &80.0 &57.1\\
			Pneumothorax & 80.0 &57.1 &66.7 & 80.0 &57.1 &66.7\\
			Consolidation & 16.3 &87.5 &27.5 & 77.8 &87.5 &82.4 \\
			Edema & 66.7 &90.9 &76.9 & 76.9 &90.9 &83.3\\
			Emphysema & 94.1 &64.0 &76.2 &94.1 &64.0 &76.2\\
			Fibrosis & 100.0 &100.0 &100.0 &100.0 &100.0 &100.0\\
			PT & 100.0 &75.0 &85.7 &100.0 &75.0 &85.7 \\
			Hernia & 100.0 &100.0 &100.0 &100.0 &100.0 &100.0\\
			\hspace*{1em}\textit{Total} & 77.2& 84.6& 80.7 & 89.8 &85.0 &87.3\\
			\hline
			\multicolumn{7}{c}{ChestX-ray14}\\
			\hline
			Atelectasis & 88.6& 98.1& 93.1 & 96.6& 97.3& 96.9\\
			Cardiomegaly & 94.1& 95.7& 94.9  & 96.7& 95.7& 96.2\\
			Effusion & 87.7& 99.6& 93.3  & 94.8& 99.2& 97.0\\
			Infiltration & 69.7& 90.0& 78.6&95.9& 85.6& 90.4\\
			Mass & 85.1& 92.5& 88.7 & 92.5& 92.5& 92.5\\
			Nodule & 78.4& 92.3& 84.8  & 84.5& 92.3& 88.2\\
			Pneumonia & 73.8& 87.3& 80.0 & 88.9& 87.3& 88.1\\
			Pneumothorax & 87.4& 100.0& 93.3 & 94.3& 98.8& 96.5\\
			Consolidation & 72.8& 98.3& 83.7 & 95.2& 98.3& 96.7 \\
			Edema &72.1& 93.9& 81.6 & 96.9& 93.9& 95.43\\
			Emphysema & 97.6& 93.2& 95.3 &100.0& 90.9& 95.2\\
			Fibrosis & 84.6& 100.0& 91.7 &91.7& 100.0& 95.7\\
			PT & 85.1& 97.6& 90.9 &97.6& 97.6& 97.6 \\
			Hernia & 66.7 &100.0 &80.0  &100.0 &100.0 &100.0\\
			\hspace*{1em}\textit{Total} & 82.8 &95.5 &88.7 & 94.4 &94.4 &94.4\\
			\hline
		\end{tabular}
	\end{center}
	\caption{Evaluation of image labeling results on OpenI and ChestX-ray14 dataset. Performance is reported using P, R, F1-score. PT: Pleural Thickening}
	\label{tab:evaluation_openi_14}
\end{table*}

\subsection{ Benchmark Results}
\begin{figure}[t]
	\centering
	\includegraphics[width=1\linewidth]{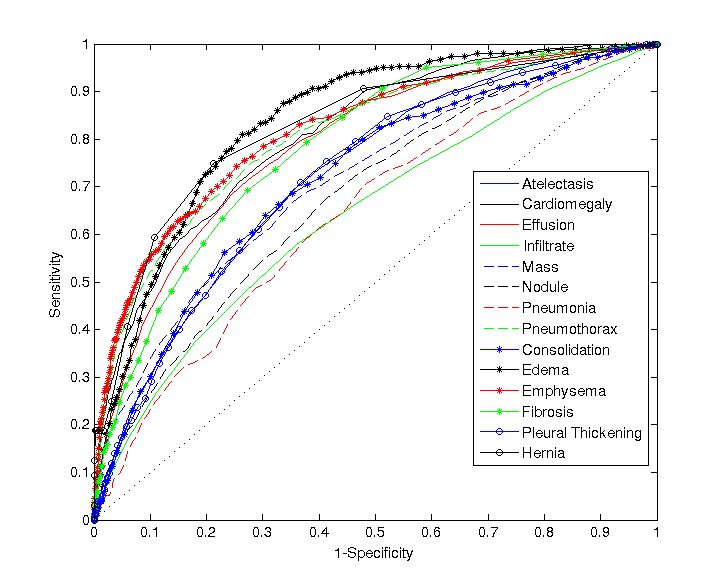} 
	\caption{Multi-label classification performance on ChestX-ray14 with ImageNet pre-trained ResNet.}
	\label{fig:ROC_14}
\end{figure}

In a similar fashion to the experiment on ChestX-ray8, we evaluate and validate the unified disease classification and localization framework on ChestX-ray14 database. In total, 112,120 frontal-view X-ray images are used, of which 51,708 images contain one or more pathologies. The remaining 60,412 images do not contain the listed 14 disease findings. For the pathology classification and localization task, we randomly shuffled the entire dataset into three subgroups on the patient level for CNN fine-tuning via Stochastic Gradient Descent (SGD): i.e. training ($\sim70\%$), validation ($\sim10\%$) and testing ($\sim20\%$). All images from the same patient will only appear in one of the three sets.  \footnote{ Data split files could be downloaded via \url{https://nihcc.app.box.com/v/ChestXray-NIHCC}} We report the 14 thoracic disease recognition performance on the published testing set  in comparison with the counterpart based on ChestX-ray8, shown in Table {\color{red} 17} and Figure \ref{fig:ROC_14}.

Since the annotated B-Boxes of pathologies are unchanged, we only test the localization performance on the original 8 categories. Results measured by the Intersection over the detected B-Box area ratio (IoBB) (similar to Area of Precision or Purity) are demonstrated in Table {\color{red} 19}. 

\begin{table*}[t]
	\small
	\centering
	\begin{tabular}{|p{6.5em}||c|c|c|c|c|c|c|c|}
		\hline
		{\bf T(IoBB)}   & {\bf Atelectasis} & {\bf Cardiomegaly} & {\bf Effusion} & {\bf Infiltration} & {\bf Mass} & {\bf Nodule} & {\bf Pneumonia} & {\bf Pneumothorax} \\
		\hline\hline
		\multicolumn{9}{|c|}{T(IoBB) = 0.1}\\
		\hline
		{\bf Acc.}   & 0.6222& 1&	0.7974&0.9106&0.5882&0.1519&0.8583&0.5204\\
		\hline
		{\bf AFP}    & 0.8293&0.1768&0.6148&0.4919&0.3933&0.4685&0.4360&0.4543\\
		\hline
		\multicolumn{9}{|c|}{\textbf{T(IoBB) = 0.25} (Two times larger on both x and y axis than ground truth B-Boxes)}\\
		\hline
		{\bf Acc.}   & 0.3944&0.9863&0.6339&0.7967&0.4588&0.0506&0.7083&0.3367\\
		\hline
		{\bf AFP}   & 0.9319&0.2042&0.6880&0.5447&0.4288&0.4786&0.4959&0.4857\\
		\hline
		\multicolumn{9}{|c|}{T(IoBB) = 0.5}\\
		\hline
		{\bf Acc.}   & 0.1944&0.9452&0.4183&0.6504&0.3058&0&0.4833&0.2653\\
		\hline
		{\bf AFP}   &0.9979&0.2785&0.7652&0.6006&0.4604&0.4827&0.5630&0.5030\\
		\hline
		\multicolumn{9}{|c|}{T(IoBB) = 0.75}\\
		\hline		
		{\bf Acc.}   & 0.0889&0.8151&0.2287&0.4390&0.1647&0&0.2917&0.1735\\
		\hline
		{\bf AFP}   & 1.0285&0.4045&0.8222&0.6697&0.4827&0.4827&0.6169&0.5243\\
		\hline
		\multicolumn{9}{|c|}{T(IoBB) = 0.9} \\
		\hline
		{\bf Acc.}   &0.0722&0.6507&0.1373&0.3577&0.0941&0&0.2333&0.1224\\
		\hline
		{\bf AFP}   & 1.0356&0.4837&0.8445&0.7043&0.4939&0.4827&0.6331&0.5346\\
		\hline
	\end{tabular}\label{tab:IoBB_14}
	\caption{Pathology localization accuracy and average false positive number for ChestX-ray14.}
\end{table*}

Overall, both of the classification and localization performance on ChestX-ray14 is equivalent to the counterpart on ChestX-ray8. 

\begin{figure*}[t]
	\centering
	\includegraphics[width=0.8\linewidth]{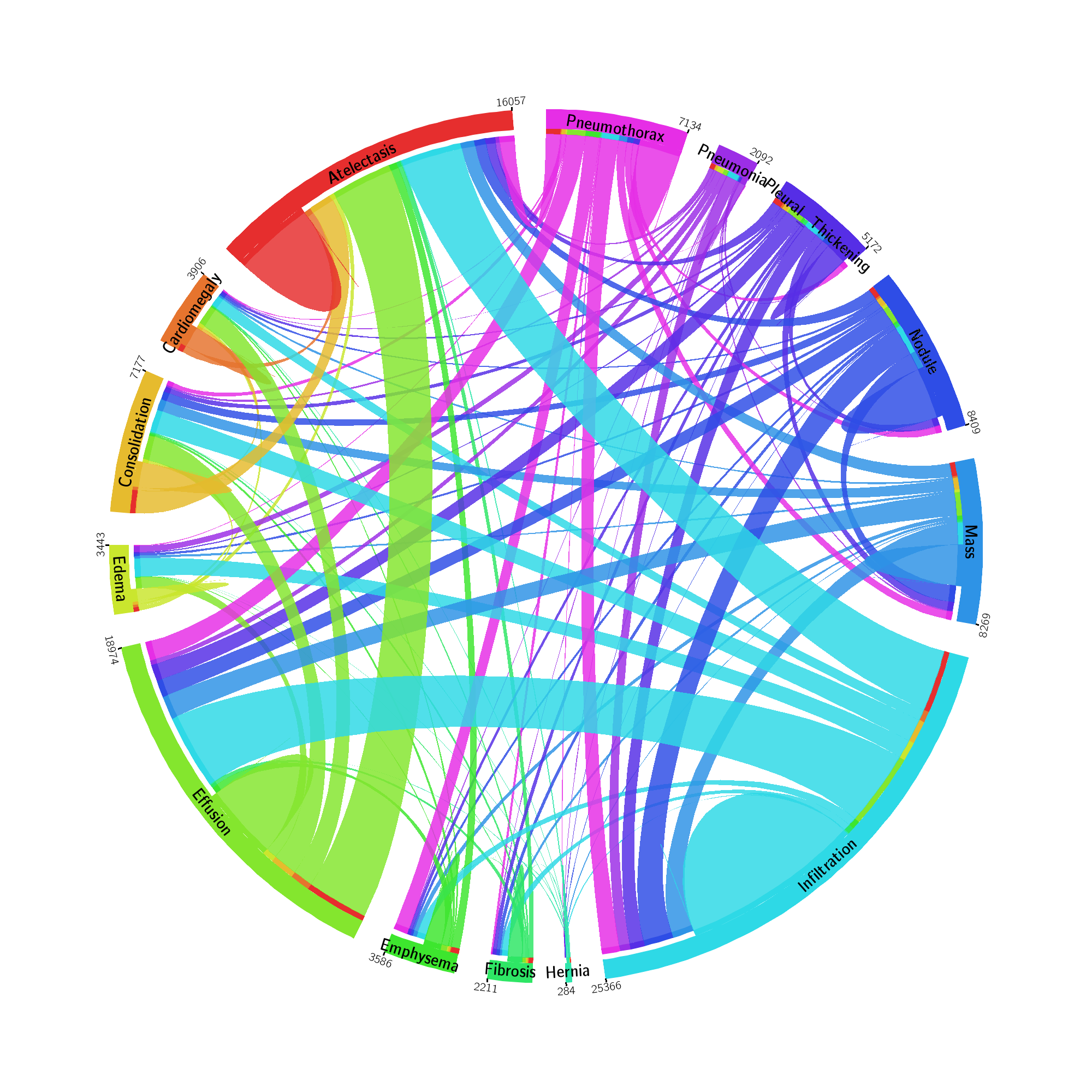}
	\caption{The circular diagram shows the proportions of images with multi-labels in each of 14 pathology classes and the labels' co-occurrence statistics.}
	\label{fig:s14}
\end{figure*}

\end{document}